%% file: arxiv.tex
\documentclass[10pt,twocolumn,letterpaper]{article}

\usepackage{cvpr}
\usepackage{times}
\usepackage{epsfig}
\usepackage{graphicx}
\usepackage{amsmath}
\usepackage{amssymb}
\usepackage[utf8]{inputenc} 
\usepackage[T1]{fontenc}    
\usepackage{url}            
\usepackage{booktabs}       
\usepackage{amsfonts}       
\usepackage{caption}
\usepackage{graphicx,epstopdf}
\usepackage{multirow}
\usepackage{float}
\usepackage{bbm}
\usepackage{enumitem}
\usepackage{colortbl}
\usepackage{xifthen}
\usepackage{nicefrac}       
\usepackage{microtype}      
\usepackage[dvipsnames]{xcolor}
\usepackage[linesnumbered,ruled]{algorithm2e}
\usepackage{subfigure}
\usepackage{makecell}
\usepackage{stmaryrd}
\usepackage{nicefrac}
\usepackage{algpseudocode}
\usepackage{relsize}
\usepackage{diagbox}
\usepackage{tabularx}

\input{spacesaver_bib.tex}

\usepackage[pagebackref=true,breaklinks=true,letterpaper=true,colorlinks,bookmarks=false]{hyperref}

\newcommand{\na}{{$\times$}}
\newcommand{\PowerTriangle}[3]{ \mathop{\vphantom{\triangle}}_{#1}\hspace{-0.17em}{\stackrel{#2}{\triangle}}_{#3}}

\newcommand{\defeq}{\raisebox{-0.15\totalheight}{$\triangleq$}}

\cvprfinalcopy 

\def\cvprPaperID{7888} 

\usepackage{stmaryrd}
\usepackage{trimclip}

\makeatletter
\DeclareRobustCommand{\shortto}{%
	\mathrel{\mathpalette\short@to\relax}%
}

\newcommand{\short@to}[2]{%
	\mkern2mu
	\clipbox{{.3\width} 0 0 0}{$\m@th#1\vphantom{+}{\shortrightarrow}$}%
}
\makeatother

\newcommand{\ssmallx}{{\hspace{-1pt}\scriptscriptstyle \mathcal{X}\hspace{.5pt}}}
\newcommand{\ssmally}{{\hspace{-1pt}\scriptscriptstyle \mathcal{Y}}}
\newcommand{\smallx}{{\scriptstyle \mathcal{X}}}
\newcommand{\smally}{{\scriptstyle \mathcal{Y}}}
\newcommand{\yset}{{\hspace{-1pt}\scriptscriptstyle {Y}}}
\newcommand{\splus}{{\hspace{-1pt}\scriptstyle {\times}\hspace{-1pt}}}

\begin{document}
	
	\title{\vspace{-3pt}Robust Learning Through Cross-Task Consistency\vspace{-0pt}}

	\author{Amir R. Zamir$^{\dagger*}$ \;\; Alexander Sax$^{\ddagger*}$ \;\; Teresa Yeo$^{\dagger}$ \;\; Oğuzhan Kar$^{\dagger}$ \;\; Nikhil Cheerla$^{\S}$  \\ Rohan Suri$^{\S}$ \;\;  Zhangjie Cao$^{\S}$ \;\;  Jitendra Malik$^{\ddagger}$ \;\; Leonidas Guibas$^{\S}$\vspace{10pt}\\ 
		$^\dagger$ Swiss Federal Institute of Technology (EPFL)  \;\;  
		$^\S$ Stanford University  \;\;  
		$^\ddagger$ UC Berkeley\vspace{10pt}\\ 
		\textcolor{blue}{\url{http://consistency.epfl.ch/}\vspace{-6pt}}
	}

	\maketitle
	
	\begin{abstract}
		Visual perception entails solving a wide set of tasks, e.g., object detection, depth estimation, etc. The predictions made for multiple tasks from the same image are not independent, and therefore, are expected to be ‘consistent’. We propose a broadly applicable and fully computational method for augmenting learning with \textbf{Cross-Task Consistency}.\footnote{Abbreviated {\textbf{X-TC}}, standing for \textbf{Cross}-\textbf{T}ask \textbf{C}onsistency.\\   \phantom{....} *Equal.}
		The proposed formulation is based on \textbf{inference-path invariance} over a graph of arbitrary tasks. We observe that learning with cross-task consistency leads to more accurate predictions and better generalization to out-of-distribution inputs. This framework also leads to an informative unsupervised quantity, called \textbf{Consistency Energy}, based on measuring the intrinsic consistency of the system. Consistency Energy correlates well with the supervised error ($r{=}0.67$), thus it can be employed as an unsupervised confidence metric as well as for detection of out-of-distribution inputs ($\text{ROC-AUC=}0.95$). The evaluations are performed on multiple datasets, including Taskonomy, Replica, CocoDoom, and ApolloScape, and they benchmark cross-task consistency versus various baselines including conventional multi-task learning, cycle consistency, and analytical consistency.

	\end{abstract}

	\spacesaversection{Introduction}
	\label{sec:intro}
	{\textbf{What}} is consistency: suppose an object detector detects a ball in a particular region of an image, while a depth estimator returns a flat surface for the same region. This presents an issue -- at least one of them has to be wrong, because they are \emph{inconsistent}. More concretely, the first prediction domain (objects) and the second prediction domain (depth) are not independent and consequently enforce some constraints on each other, often referred to as \emph{consistency constraints}.
	
	\begin{figure}
		\includegraphics[trim={0mm 67mm 0mm 0},clip,width=1\columnwidth]{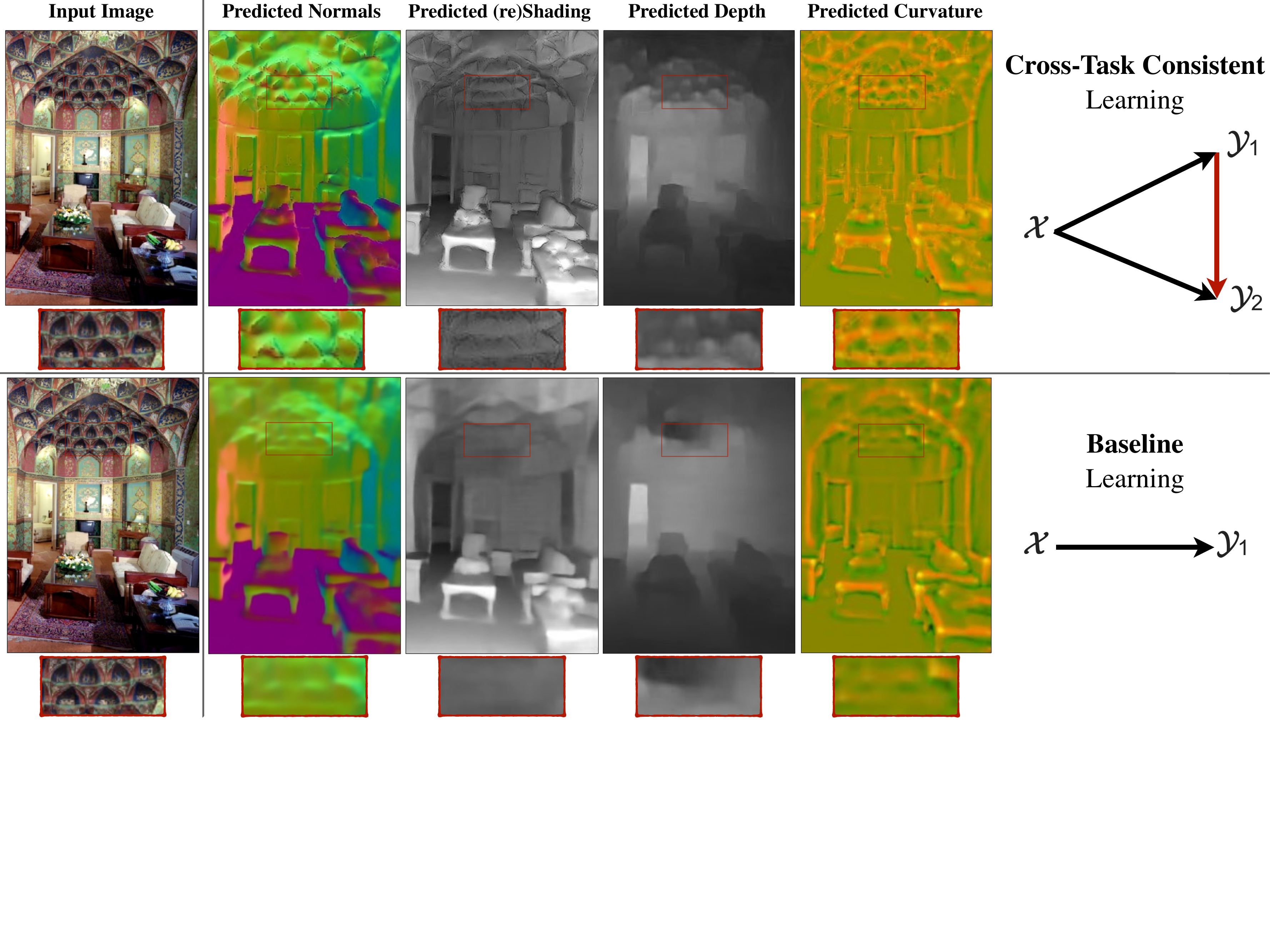}
		\vspace{-5mm}
		\caption{\footnotesize{\textbf{Cross-Task Consistent Learning.}} The predictions made for different tasks out of one image are expected to be \emph{consistent}, as the underlying scene is the same. This is exemplified by a challenging query and four sample predictions out of it. We propose a general method for learning utilizing data-driven cross-task consistency constraints. The lower and upper rows show the results of the baseline (independent learning) and learning with consistency, which yields higher quality and more consistent predictions. Red boxes provide magnifications. [Best seen on screen]}
		\label{fig:motiv}
	\end{figure}
	
	{\textbf{Why}} is it important to incorporate consistency in learning: first, desired learning tasks are usually predictions of different aspects of one underlying reality (the scene that underlies an image). Hence inconsistency among predictions implies contradiction and is inherently undesirable. Second, consistency constraints are informative and can be used to better fit the data or lower the sample complexity. Also, they may reduce the tendency of neural networks to learn ``surface statistics'' (superficial cues)~\cite{jo2017measuring}, by enforcing constraints rooted in different physical or geometric rules. This is empirically supported by the improved generalization of models when trained with consistency constraints (Sec.~\ref{sec:results}).
	
\begin{figure}
		\includegraphics[trim={0mm 206mm 0mm 0},clip,width=1\columnwidth]{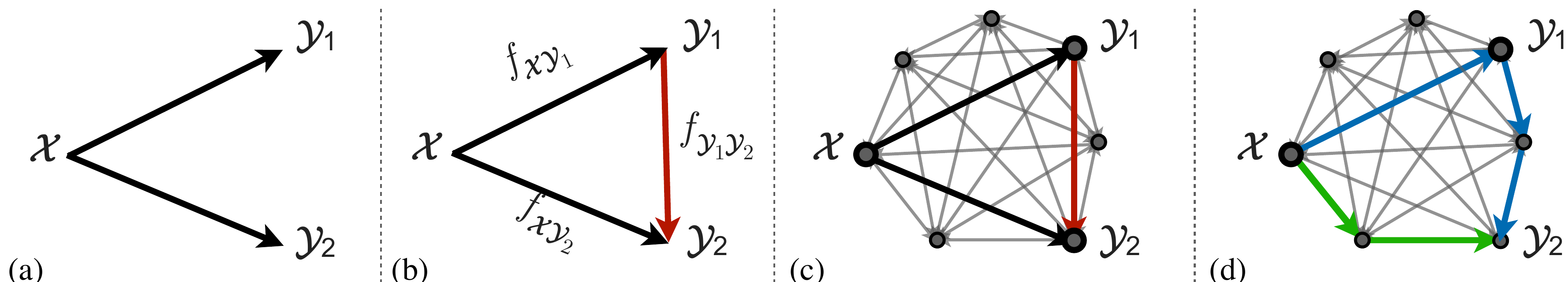}
		\vspace{-7mm}
		\caption{\footnotesize{\textbf{Enforcing Cross-Task Consistency:}
				\textbf{(a)} shows the typical multitask setup where predictions $\smallx{\shortrightarrow}\smally_1$ and $\smallx{\shortrightarrow}\smally_2$ are trained without a notation of consistency. \textbf{(b)} depicts the elementary \emph{triangle consistency constraint} where the prediction $\smallx{\shortrightarrow}\smally_1$ is enforced to be consistent with $\smallx{\shortrightarrow}\smally_2$ using a function that relates $\smally_1$ to $\smally_2$ (i.e. $\smally_1 \textcolor{red}{\shortrightarrow}\smally_2$). \textbf{(c)} shows how the triangle unit from (b) can be an element of a larger system of domains. Finally, \textbf{(d)} illustrates the generalized case where in the larger system of domains, consistency can be enforced using invariance along arbitrary paths, as long as their endpoints are the same (here the blue and green paths). This is the general concept behind \emph{inference-path invariance}. The triangle in (b) is the smallest unit of such paths.}}
		\label{fig:graph}
	\end{figure}

	{\textbf{How}} can we design a learning system that makes consistent predictions: this paper proposes a method which, given an arbitrary dictionary of tasks, augments the learning objective with explicit constraints for cross-task consistency. The constraints are learned from data rather than apriori given relationships.\footnote{For instance, it is not necessary to encode that surface normals are the 3D derivative of depth or occlusion edges are discontinuities in depth.} This makes the method applicable to any pairs of tasks as long as they are not statistically independent; \emph{even if their analytical relationship is unknown, hard to program, or non-differentiable}. 
	The primary concept behind the method is `inference-path invariance'. 
	That is, the result of inferring an \textcolor{brown}{output} domain from an \textcolor{teal}{input} domain should be the same, regardless of the \textcolor{purple}{intermediate} domains mediating the inference (e.g., \textcolor{teal}{RGB}$\shortrightarrow$\textcolor{brown}{normals} and \textcolor{teal}{RGB}$\shortrightarrow$\textcolor{purple}{depth}$\shortrightarrow$\textcolor{brown}{normals} and \textcolor{teal}{RGB}$\shortrightarrow$\textcolor{purple}{shading}$\shortrightarrow$\textcolor{brown}{normals} are expected to yield the same normals result). 
	When inference paths with the same endpoints, but different intermediate domains, yield similar results, this implies the intermediate domain predictions did not conflict as far as the output was concerned. 
	We apply this concept over paths in a graph of tasks, where the nodes and edges are prediction domains and neural network mappings between them, respectively (Fig.~\ref{fig:graph}(d)).   
	Satisfying this invariance constraint over \emph{all} paths in the graph ensures the predictions for all domains are in global cross-task agreement.\footnote{inference-path invariance was inspired by \textbf{Conservative Vector Fields} in vector calculus and physics that are (at a high level) fields in which integration along \emph{different paths yield the same results, as long as their endpoints are the same}~\cite{guillemin1974differential}. Many key concepts in physics are `conservative', e.g., gravitational force: the work done against gravity when moving between two points is independent of the path taken.} 
	
	To make the associated large optimization job manageable, we reduce the problem to a `separable' one, devise a tractable training schedule, and use a `perceptual loss' based formulation. The last enables mitigating residual errors in networks and potential ill-posed/one-to-many mappings between domains; this is crucial as one may not be able to always infer one domain from another with certainty (Sec.~\ref{sec:method}).

	\href{https://consistency.epfl.ch/visuals}{Interactive visualizations}, \href{https://consistency.epfl.ch/#models}{trained models}, \href{https://consistency.epfl.ch/#models}{code}, and a \href{https://consistency.epfl.ch/demo}{live demo} are available at \href{http://consistency.epfl.ch/}{http://consistency.epfl.ch/}.

	
	\spacesaversection{Related Work} 
	The concept of consistency and methods for enforcing it are related to various topics, including structured prediction, graphical models~\cite{Koller:2009:PGM:1795555}, functional maps~\cite{Ovsjanikov:2012:FMF:2185520.2185526}, and certain topics in vector calculus and differential topology~\cite{guillemin1974differential}. We review the most relevant ones in context of computer vision. 
	
	\textbf{Utilizing consistency:} Various consistency constraints have been commonly found beneficial across different fields, e.g., in language as `back-translation'~\cite{BrislinCrossCultural, Artetxe18Multilingual, Lample19crosslingual, backtranslation2018} or in vision over the temporal domain~\cite{WangCycleConsistency19,DwibediTimeConsistency19}, 3D geometry \cite{godard2017unsupervised,Geonet18,garg2016unsupervised,hickson2019floors,ZhouKAHE16,zhang2017physically,Huang:2013:CSM:2600289.2600314,yin2019enforcing,zou2018df,Zhang18PathInvariant,kusupati2019normal,cosmo2017consistent}, and in recognition and (conditional/unconditional) image translation~\cite{hertzmann2001image,mirza2014conditional,pix2pix,cycleGan17,HoffmancyCADA17,choi2018stargan}. In computer vision, consistency has been extensively utilized in the cycle form and often between two or few domains~\cite{cycleGan17,HoffmancyCADA17}. In contrast, we consider consistency in the more general form of arbitrary paths with varied-lengths over a large task set, rather than the special cases of short cyclic paths. Also, the proposed approach needs \emph{no prior explicit knowledge} about task relationships~\cite{Geonet18,kusupati2019normal,yin2019enforcing,zou2018df}. 
	
	\textbf{Multi-task learning:} In the most conventional form, multi-task learning predicts multiple output domains out of a shared encoder/representation for an input. It has been speculated that the predictions of a multi-task network may be automatically cross-task consistent as the representation from which the predictions are made are shared. This has been observed to not be necessarily true in several works~\cite{Kokkinos16,Zhang17MTL,xu2018pad,standley2019}, as consistency is not directly enforced during training. We also make the same observation (see visuals \href{http://consistency.epfl.ch/visuals/}{here}) and quantify it (see Fig.~\ref{fig:energy_over_time}), which signifies the need for explicit augmentation of consistency in learning. 
	
	\textbf{Transfer learning} predicts the output of a target task given another task's solution as a source. The predictions made using transfer learning are sometimes assumed to be cross-task consistent, which is often found to not be the case~\cite{taskonomy18,sharif2014cnn}, as transfer learning does not have a specific mechanism to impose consistency by default. Unlike basic multi-task learning and transfer learning, the proposed method includes explicit mechanisms for learning with general data-driven consistency constraints. 
	
	\textbf{Uncertainty metrics:}
	Among the existing approaches to measuring prediction uncertainty, the proposed Consistency Energy (Sec.~\ref{sec:energy_method}) is most related to Ensemble Averaging~\cite{lakshminarayanan2017simple}, with the key difference that the estimations in our ensemble are from \emph{different cues/paths}, rather than retraining/reevaluating the same network with different random initializations or parameters. Using multiple cues is expected to make the ensemble more effective at capturing uncertainty.

	\begin{figure*}
		\vspace{-3mm}	
		\includegraphics[trim={0mm 208mm 0mm 0},clip,width=1\textwidth]{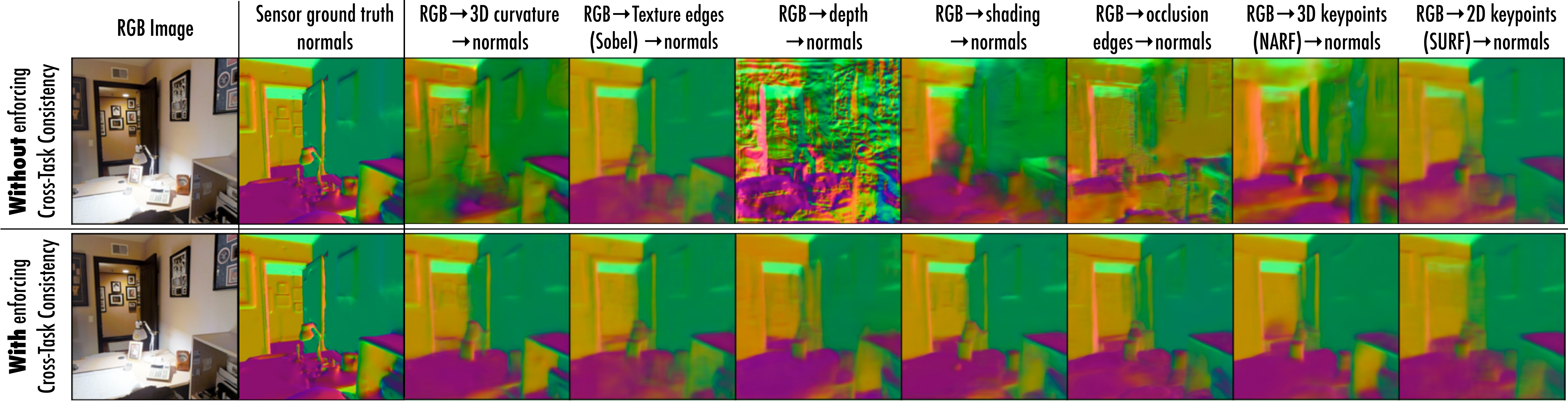}
		\vspace{-7mm}		
		\caption{\footnotesize{\textbf{Impact of disregarding cross-task consistency in learning}, illustrated using surface normals domain. Each subfigure shows the results of predicting surface normals out of the prediction of an intermediate domain; using the notation $\smallx{\shortrightarrow}\smally_1\textcolor{red}{\shortrightarrow}\smally_2$, here $\smallx$ is RGB image, $\smally_2$ is surface normals, and each column represents a different $\smally_1$. The \textbf{upper row} demonstrates the normals are noisy and dissimilar when cross-task consistency is not incorporated in learning of $\smallx{\shortrightarrow}\smally_1$ networks. Whereas enforcing consistency when learning $\smallx{\shortrightarrow}\smally_1$ results in more consistent and better normals (the \textbf{lower row}). We will show this causes the predictions for the intermediate domains themselves to be more accurate and consistent. More examples available in \href{http://consistency.epfl.ch/supplementary_material}{supplementary material}.} The \emph{Consistency Energy} (Sec.~\ref{sec:energy_method}) captures the variance among predictions in each row. }
		\label{fig:qualitative_consistency}
		\vspace{-3mm}		
	\end{figure*}

	\vspace{-0pt}
	\spacesaversection{Method}
	\label{sec:method}
	\vspace{-2pt}	
	We define the problem as follows: suppose $\smallx$ denotes the query domain (e.g., RGB images) and $\smally{=}\{\smally_1,$...$,\smally_n\}$ is the set of $n$ desired prediction domains (e.g., normals, depth, objects, etc). An individual datapoint from domains $(\smallx,\smally_1,$...$,\smally_n)$ is denoted by $(x,y_1,$...$,y_n)$. The goal is to learn functions that map the query domain onto the prediction domains, i.e. $\mathcal{F}_\ssmallx {=} \{f_{\ssmallx\ssmally_j} | \smally_j {\in} \smally \}$ where $f_{\ssmallx\ssmally_j}(x)$ outputs $y_j$ given $x$. 
	We also define $\mathcal{F}_\ssmally {=} \{f_{\ssmally_i\ssmally_j} | \smally_i, \smally_j {\in} \smally, i{\neq}j \}$, which is the set of `cross-task' functions that map the prediction domains onto each other; we use them in the consistency constraints. For now assume  $\mathcal{F}_\ssmally$ is given apriori and frozen; in Sec.~\ref{sec:full_graph} we discuss all functions $f$s are neural networks in this paper, and we learn $\mathcal{F}_\ssmally$ just like $\mathcal{F}_\ssmallx$.

	\subsection{Triangle: The Elementary Consistency Unit}
	\label{sec:triangle}
	The typical supervised way of training the neural networks in $\mathcal{F}_\ssmallx$, e.g., $f_{\ssmallx\ssmally_1}(x)$, is to find parameters of $f_{\ssmallx\ssmally_1}$ that minimize a loss with the general form $|f_{\ssmallx\ssmally_1}(x) \text{-} {y}_1|$ using a distance function as $|.|$, e.g., $\ell_1$ norm.
	This standard \emph{independent} learning of $f_{\ssmallx\ssmally_i}$s satisfies various desirable properties, including cross-task consistency, if given infinite amount of data, but not under the practical finite data regime. This is qualitatively illustrated in Fig.~\ref{fig:qualitative_consistency} (upper). Thus we introduce additional constraints to guide the training toward cross-task consistency.   
	We define the loss for predicting domain $\smally_1$ from $\smallx$ \emph{while enforcing consistency with domain $\smally_2$} as a directed triangle depicted in Fig.~\ref{fig:graph}(b):        
	\begin{equation}\hspace{-3pt}
	\mathcal{L}_{\ssmallx\ssmally_1\ssmally_2}^{\textit{\tiny{triangle}}} \defeq 
	|f_{\ssmallx\ssmally_1}\hspace{-2pt}(x) \text{-} {y}_1| \text{+} |f_{\ssmally_1\ssmally_2}{\circ}f_{\ssmallx\ssmally_1}\hspace{-2pt}(x) \text{-} f_{\ssmallx\ssmally_2}\hspace{-2pt}(x)|  
	\text{+} |f_{\ssmallx\ssmally_2}\hspace{-2pt}(x) \text{-} {y}_2|.
	\label{eq:triangle}
	\end{equation}
	The first and last terms are the standard \emph{direct} losses for training $f_{\ssmallx\ssmally_1}$ and $f_{\ssmallx\ssmally_2}$. The middle term is the \emph{consistency term} which enforces that predicting $\smally_2$ out of the predicted $\smally_1$ yields the same result as directly predicting $\smally_2$ out of $\smallx$ (done via the given cross-task function $f_{\ssmally_1\ssmally_2}$).\footnote{Operator ${\circ}$ denotes function composition: $g{\circ}h(x){\defeq}g(h(x))$.} Thus learning to predict $\smally_1$ and $\smally_2$ are not independent anymore. 
	
	The triangle loss~\ref{eq:triangle} is the smallest unit of enforcing cross-task consistency. Below we make two improving modifications on it via function `separability' and `perceptual losses'.
	
	\subsubsection{Separability of Optimization Parameters}
	The loss $\mathcal{L}_{\ssmallx\ssmally_1\ssmally_2}^{\textit{\tiny{triangle}}}$ involves \emph{simultaneous} training of two networks $f_{\ssmallx\ssmally_1}$ and $f_{\ssmallx\ssmally_2}$, thus it is resource demanding. We show $\mathcal{L}_{\ssmallx\ssmally_1\ssmally_2}^{\textit{\tiny{triangle}}}$ can be reduced to a `separable' function~\cite{stewart2012essential} resulting in two terms that can be optimized independently. 
	
	From the triangle inequality we can derive:
	\begin{equation*}
	|f_{\ssmally_1\ssmally_2}{\circ}f_{\ssmallx\ssmally_1}\hspace{-1pt}(x) \text{-} f_{\ssmallx\ssmally_2}\hspace{-1pt}(x)|{\leq}|f_{\ssmally_1\ssmally_2}{\circ}f_{\ssmallx\ssmally_1}\hspace{-1pt}(x) \text{-} {y}_2| \text{+} |f_{\ssmallx\ssmally_2}\hspace{-1pt}(x) \text{-} {y}_2|,
	\label{eq:separableineq}
	\end{equation*}
	which after	substitution in Eq.~\ref{eq:triangle} yields:
	\begin{equation}
	\mathcal{L}_{\ssmallx\ssmally_1\ssmally_2}^{\textit{\tiny{triangle}}} {\leq} 
	|f_{\ssmallx\ssmally_1}(x) \text{-} {y}_1| \text{+} |f_{\ssmally_1\ssmally_2}{\circ}f_{\ssmallx\ssmally_1}(x) \text{-} {y}_2|  
	\text{+} 2|f_{\ssmallx\ssmally_2}(x) \text{-} {y}_2|.
	\label{eq:separableineqtriangle}
	\end{equation}        
	The upper bound for $\mathcal{L}_{\ssmallx\ssmally_1\ssmally_2}^{\textit{\tiny{triangle}}}$ in inequality~\ref{eq:separableineqtriangle} can be optimized in lieu of $\mathcal{L}_{\ssmallx\ssmally_1\ssmally_2}^{\textit{\tiny{triangle}}}$ itself, as they both have the same minimizer.\footnote{Both sides of inequality~\ref{eq:separableineqtriangle} are ${\geq}0$ and $\text{=}0$ for the minimizer $f_{\ssmallx\ssmally_1}(x)\text{=} {y}_1$ \& $f_{\ssmallx\ssmally_2}(x)\text{=} {y}_2$.} 
	The terms of this bound include either $f_{\ssmallx\ssmally_1}$ or $f_{\ssmallx\ssmally_2}$, but not both, hence we now have a loss separable into functions of $f_{\ssmallx\ssmally_1}$ or $f_{\ssmallx\ssmally_2}$, and they can be optimized independently. The part pertinent to the network $f_{\ssmallx\ssmally_1}$ is:
	\begin{equation}
	\mathcal{L}_{\ssmallx\ssmally_1\ssmally_2}^{\textit{\tiny{separate}}} \defeq |f_{\ssmallx\ssmally_1}(x) - {y}_1| + |f_{\ssmally_1\ssmally_2}{\circ}f_{\ssmallx\ssmally_1}(x) - {y}_2|, 
	\label{eq:pathloss}
	\end{equation}   
	named \emph{separate}, as we reduced the closed triangle objective \scriptsize$\PowerTriangle{\smallx}{\smally_1}{\smally_2}$\normalsize in Eq.~\ref{eq:triangle} to two separate path objectives {${\smallx}{\shortrightarrow}{\smally_1}{\shortrightarrow}{\smally_2}$} and {${\smallx}{\shortrightarrow}{\smally_2}$}. The first term of Eq.~\ref{eq:pathloss} enforces the general correctness of predicting $\smally_1$, and the second term enforces its consistency with $\smally_2$ domain.

	\begin{figure*}
		\vspace{-4mm}
		\centering
		\includegraphics[trim={3mm 30mm 71mm 108mm},clip,width=1\textwidth]{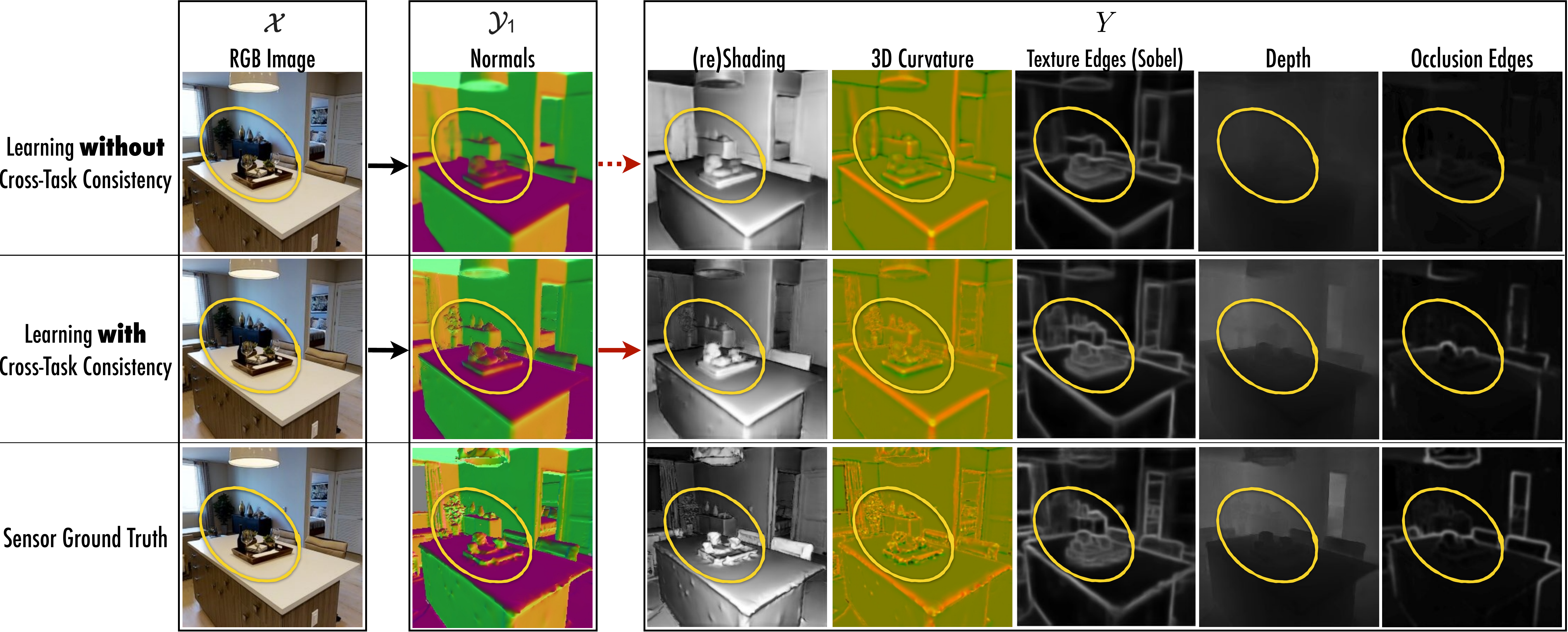}
		\vspace{-7mm}
		\caption{\footnotesize{\textbf{Learning with and without cross-task consistency shown for a sample query}. Using the notation $\smallx{\shortrightarrow}\smally_1\textcolor{red}{\shortrightarrow}Y$, here $\smallx$ is RGB image, $\smally_1$ is surface normals, and five domains in $Y$ are reshading, 3D curvature, texture edges (Sobel filter), depth, and occlusion edges. 
				\\ \textbf{Top row} shows the results of standard training of $\smallx{\shortrightarrow}\smally_{1}$. After convergence of training, the predicted normals ($\smally_1$) are projected onto other domains ($Y$) which reveal various inaccuracies. This demonstrates such cross-task projections $\smally_1 \textcolor{red}{\shortrightarrow} Y$ can provide additional cues to training $\smallx{\shortrightarrow}\smally_1$.
				\\ \textbf{Middle row} shows the results of consistent training of $\smallx{\shortrightarrow}\smally_1$ by leveraging $\smally_1{\textcolor{red}{\shortrightarrow}}Y$ in the loss. The predicted normals are notably improved, especially in hard to predict \emph{fine-grained details} (zoom into the yellow markers. Best seen on screen). 
				\\ \textbf{Bottom row} provides the ground truth. See video examples at \href{https://consistency.epfl.ch/visuals}{visualizations webpage}.}}
		\label{fig:qualitative_method}
		\vspace{-3mm}
	\end{figure*}

	\subsubsection{Reconfiguration into a ``Perceptual Loss''}   
	\label{sec:perceploss}
	\label{sec:sep_paired} 
	Training $f_{\ssmallx\ssmally_1}$ using the loss $\mathcal{L}_{\ssmallx\ssmally_1\ssmally_2}^{\textit{\tiny{separate}}}$ requires a training dataset with multi domain annotations for one input: $(x,{y_1},{y_2})$. It also relies on availability of a \emph{perfect} function $f_{\ssmally_1\ssmally_2}$ for mapping $\smally_1$ onto $\smally_2$; i.e. it demands ${y_2}{=}f_{\ssmally_1\ssmally_2}({y_1})$. We show how these two requirements can be reduced.  
	
	Again, from triangle inequality we can derive:
	\begin{multline}
	|f_{\ssmally_1\ssmally_2}{\circ}f_{\ssmallx\ssmally_1}(x) - {y}_2| {\leq}  |f_{\ssmally_1\ssmally_2}{\circ}f_{\ssmallx\ssmally_1}(x) - f_{\ssmally_1\ssmally_2}({y}_1)| + \\ |f_{\ssmally_1\ssmally_2}({y}_1) - {y}_2|,
	\end{multline}        
	which after substitution in Eq.~\ref{eq:pathloss} yields:        
	\begin{multline}
	\mathcal{L}_{\ssmallx\ssmally_1\ssmally_2}^{\textit{\tiny{separate}}} {\leq} |f_{\ssmallx\ssmally_1}(x) - {y}_1| +  |f_{\ssmally_1\ssmally_2}{\circ}f_{\ssmallx\ssmally_1}(x) - f_{\ssmally_1\ssmally_2}({y}_1)| + \\ |f_{\ssmally_1\ssmally_2}({y}_1) - {y}_2|.
	\label{eq:percepinequality} 
	\end{multline}           
	Similar to the discussion for inequality~\ref{eq:separableineqtriangle}, the upper bound in inequality~\ref{eq:percepinequality} can be optimized in lieu of $\mathcal{L}_{\ssmallx\ssmally_1\ssmally_2}^{\textit{\tiny{separate}}}$ as both have the same minimizer.\footnote{Both sides of inequality \ref{eq:percepinequality} are ${\geq}0$ and ${=}0$ for the minimizer $f_{\ssmallx\ssmally_1}(x){=} {y}_1$. The term $|f_{\ssmally_1\ssmally_2}({y}_1) - {y}_2|$ is a constant and ${\sim}0$, as it is exactly the training objective of $f_{\ssmally_1\ssmally_2}$. The non-zero residual should be ignored and assumed 0 as the non-zero part is irrelevant to $f_{\ssmallx\ssmally_1}$, but imperfections of $f_{\ssmally_1\ssmally_2}$.}
	As the last term is a constant w.r.t. $f_{\ssmallx\ssmally_1}$, the final loss for training $f_{\ssmallx\ssmally_1}$ subject to consistency with domain $\smally_2$ is:       
	\begin{equation}
	\mathcal{L}_{\ssmallx\ssmally_1\ssmally_2}^{\textit{\tiny{perceptual}}} \defeq |f_{\ssmallx\ssmally_1}(x) - {y}_1| + |f_{\ssmally_1\ssmally_2}{\circ}f_{\ssmallx\ssmally_1}(x) - f_{\ssmally_1\ssmally_2}({y}_1)|. 
	\label{eq:percep_loss}
	\end{equation}           
	The loss $\mathcal{L}_{\ssmallx\ssmally_1\ssmally_2}^{\textit{\tiny{perceptual}}}$ no longer includes ${y_2}$, hence it admits pair training data $(x,{y_1})$ rather than triplet $(x,{y_1},{y_2})$.\footnote{Generally for $n$ domains, this formulation allows using datasets of \emph{pairs} among $n$ domains, rather than one \emph{$n$-tuple} multi annotated dataset.}
	Comparing $\mathcal{L}_{\ssmallx\ssmally_1\ssmally_2}^{\textit{\tiny{perceptual}}}$ and $\mathcal{L}_{\ssmallx\ssmally_1\ssmally_2}^{\textit{\tiny{separate}}}$ shows the modification boiled down to replacing ${y_2}$ with $f_{\ssmally_1\ssmally_2}({{y_1}})$. This makes intuitive sense too, as ${y_2}$ is the match of ${y_1}$ in the $\smally_2$ domain.

	\textbf{Ill-posed tasks and imperfect networks:} If $f_{\ssmally_1\ssmally_2}$ is a \emph{noisy} estimator, then $f_{\ssmally_1\ssmally_2}({y_1}){=}{y_2{+}{noise}}$ rather than $f_{\ssmally_1\ssmally_2}({y_1}){=}{y_2}$. Using a noisy $f_{\ssmally_1\ssmally_2}$ in $\mathcal{L}_{\ssmallx\ssmally_1\ssmally_2}^{\textit{\tiny{separate}}}$ corrupts the training of $f_{\ssmallx\ssmally_1}$ since the second loss term does not reach 0 if $f_{\ssmallx\ssmally_1}(x)$ correctly outputs ${y_1}$. That is in contrast to $\mathcal{L}_{\ssmallx\ssmally_1\ssmally_2}^{\textit{\tiny{perceptual}}}$ where both terms have the same global minimum and are always 0 if $f_{\ssmallx\ssmally_1}(x)$ outputs ${y_1}$ -- even when $f_{\ssmally_1\ssmally_2}({y_1}){=}{y_2{+}{noise}}$. Thus $\mathcal{L}_{\ssmallx\ssmally_1\ssmally_2}^{\textit{\tiny{perceptual}}}$ enables a robust training of $f_{\ssmallx\ssmally_1}(x)$ w.r.t. imperfections in $f_{\ssmally_1\ssmally_2}$. This is crucial since neural networks are almost never perfect estimators, e.g., due to lacking an optimal training process for them or potential ill-posedness of the task $y_1{\shortto}y_2$. Further discussion and experiments are available in \href{http://consistency.epfl.ch/supplementary_material}{supplementary material}.
	
	\textbf{Perceptual Loss:} The process that led to Eq.~\ref{eq:percep_loss} can be generally seen as using the loss $|g{\circ}f(x){-}g(y)|$ instead of $|f(x){-}y|$. The latter compares $f(x)$ and $y$ in their explicit space, while the former compares them via the lens of function $g$. This is often referred to as ``perceptual loss'' in super-resolution and style transfer literature~\cite{DBLP:journals/corr/JohnsonAL16}--where two images are compared in the \emph{representation space} of a network pretrained on ImageNet, rather than in \emph{pixel space}. 
	Similarly, the consistency constraint between the domains $\smally_1$ and $\smally_2$ in Eq.~\ref{eq:percep_loss} (second term) can be viewed as judging the prediction $f_{\ssmallx\ssmally_1}(x)$ against ${y_1}$ via the lens of the network $f_{\ssmally_1\ssmally_2}$; here $f_{\ssmally_1\ssmally_2}$ is a ``perceptual loss'' for training $f_{\ssmallx\ssmally_1}$. However, unlike the ImageNet-based perceptual loss~\cite{DBLP:journals/corr/JohnsonAL16}, this function has the specific and interpretable job of enforcing consistency with another task. We also use multiple  $f_{\ssmally_1\ssmally_i}$s simultaneously which enforces consistency of predicting $\smally_1$  against multiple other domains (Sections \ref{sec:multitriangle} and \ref{sec:full_graph}).

	\subsection{Consistency of $f_{\ssmallx\ssmally_1}$ with `Multiple' Domains}       
	\label{sec:multitriangle} 
\begin{figure}
		\vspace{-1mm}		
		\includegraphics[trim={0mm 209mm 0mm 0},clip,width=1\columnwidth]{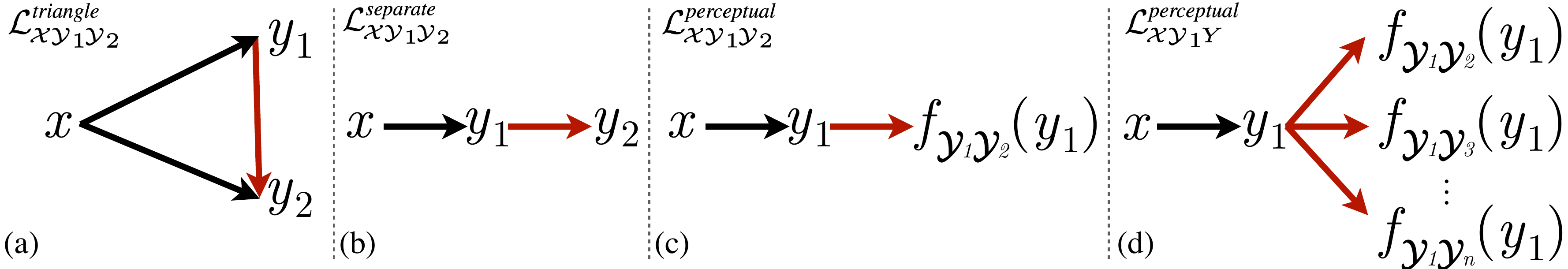}
		\vspace{-6mm}
		\caption{\footnotesize{\textbf{Schematic summary of derived losses for $f_{\ssmallx\ssmally_1}$}.\textbf{(a)}: $\mathcal{L}_{\ssmallx\ssmally_1\ssmally_2}^{\textit{\tiny{triangle}}}$ (Eq.\ref{eq:triangle}). 
				\textbf{(b)}: $\mathcal{L}_{\ssmallx\ssmally_1\ssmally_2}^{\textit{\tiny{separate}}}$ (Eq.\ref{eq:pathloss}). 
				\textbf{(c)}: $\mathcal{L}_{\ssmallx\ssmally_1\ssmally_2}^{\textit{\tiny{perceptual}}}$ (Eq.\ref{eq:percep_loss}). 
				\textbf{(d)}: $\mathcal{L}_{\ssmallx\ssmally_1{\yset}}^{\textit{\tiny{perceptual}}}$ (Eq.\ref{eq:setloss}).}		\vspace{-1mm}}
		\label{fig:all_losses}
	\end{figure}

	The derived $\mathcal{L}_{\ssmallx\ssmally_1\ssmally_2}^{\textit{\tiny{perceptual}}}$ loss augments learning of $f_{\ssmallx\ssmally_1}$ with a consistency constraint against \emph{one} domain $\smally_2$. Straightforward extension of the same derivation to enforcing consistency of $f_{\ssmallx\ssmally_1}$ against \emph{multiple} other domains (i.e. when $f_{\ssmallx\ssmally_1}$ is part of multiple simultaneous triangles) yields: 
	\begin{equation}
	\hspace{-5.5pt}\mathcal{L}_{\ssmallx\ssmally_1{\yset}}^{\textit{\tiny{perceptual}}} \defeq |Y|{\splus}|f_{\ssmallx\ssmally_1}\hspace{-2.5pt}(x) \text{-} {y}_1|\text{+}\hspace{-5pt}\sum_{\ssmally_i \in Y}\hspace{-3pt} |f_{\ssmally_1\ssmally_i}\hspace{-1.5pt}{\circ}\hspace{-1.5pt}f_{\ssmallx\ssmally_1}\hspace{-2.5pt}(x) \text{-} f_{\ssmally_1\ssmally_i}\hspace{-2pt}({y}_1)|, 
	\hspace{-2.5pt}\label{eq:setloss}
	\end{equation}    
	where $Y$ is the \emph{set} of domains with which $f_{\ssmallx\ssmally_1}$ must be consistent, and $|Y|$ is the cardinality of $Y$. Notice that $\mathcal{L}_{\ssmallx\ssmally_1\ssmally_2}^{\textit{\tiny{perceptual}}}$ is a special case of $\mathcal{L}_{\ssmallx\ssmally_1{\yset}}^{\textit{\tiny{perceptual}}}$ where $Y{=}\{{\smally_2}\}$. Fig.~\ref{fig:all_losses} summarizes the derivation of losses for $f_{\ssmallx\ssmally_1}$.
	
	Fig.~\ref{fig:qualitative_method} shows qualitative results of learning $f_{\ssmallx\ssmally_1}$ with and without cross-task consistency for a sample query.

	\subsection{Beyond Triangles: Globally Consistent  Graphs}		
	\label{sec:full_graph}
	The discussion so far provided the loss for the cross-task consistent training of \emph{one} function $f_{\ssmallx\ssmally_1}$ using elementary \emph{triangle} based units. We also assumed the functions $\mathcal{F}_\ssmally$ were given apriori. The more general multi-task setup is: given a \emph{large set of domains}, we are interested in learning functions that map the domains onto each other in a \emph{globally cross-task consistent} manner. This objective can be formulated over a graph $\mathcal{G} {=} (\mathcal{D},\mathcal{F})$ with nodes representing all of the domains  $\mathcal{D} {=} (\smallx \cup \smally$) and edges being neural networks between them $\mathcal{F} {=} (\mathcal{F}_\smallx \cup \mathcal{F}_\smally)$; see Fig.\ref{fig:graph}(c). 
	
	\textbf{Extension to Arbitrary Paths}: The transition from three domains to a large graph $\mathcal{G}$ enables forming more general consistency constraints using \emph{arbitrary-paths}. That is, two \emph{paths} with same endpoint should yield the same results -- an example is shown in Fig.\ref{fig:graph}(d). The triangle constraint in Fig.\ref{fig:graph}(b,c) is a special case of the more general constraint in Fig.\ref{fig:graph}(d), if paths with lengths 1 and 2 are picked for the green and blue paths. Extending the derivations done for a triangle in Sec.~\ref{sec:triangle} to paths yields:
	\vspace{-3mm}
	\begin{multline}
	\mathcal{L}_{\ssmallx\ssmally_1\ssmally_2...\ssmally_k}^{\textit{\tiny{perceptual}}} = |f_{\ssmallx\ssmally_1}(x) {-} {y}_1| + \\ 
	|f_{\ssmally_{k-1}\ssmally_k}{\circ}...{\circ}f_{\ssmally_1\ssmally_2}{\circ}f_{\ssmallx\ssmally_1}(x) {-} f_{\ssmally_{k-1}\ssmally_k}{\circ}...{\circ}f_{\ssmally_1\ssmally_2}({y}_1)|, 
	\label{eq:longpathloss}
	\end{multline}
	which is the loss for training $f_{\ssmallx\ssmally_1}$ using the arbitrary consistency path $\smallx{\shortrightarrow}\smally_1{\shortrightarrow}\smally_2...{\shortrightarrow}\smally_k$ with length $k$ (full derivation provided in~\href{http://consistency.epfl.ch/supplementary_material}{supplementary material}). Notice that Eq.~\ref{eq:percep_loss} is a special case of Eq.~\ref{eq:longpathloss} if $k{=}2$. Equation~\ref{eq:longpathloss} is particularly useful for incomplete graphs; if the function $\smally_1{\shortrightarrow}\smally_k$ is missing, consistency between domains $\smally_1$ and $\smally_k$ can still be enforced via transitivity through other domains using Eq.~\ref{eq:longpathloss}.
	
	Also, extending Eq.~\ref{eq:longpathloss} to \emph{multiple simultaneous paths} (as in Eq.~\ref{eq:setloss}) by summing the path constraints is straightforward. 
	
	\textbf{Global Consistency Objective:} We define reaching \emph{global} cross-task consistency for graph $\mathcal{G}$ as satisfying the consistency constraint for \emph{all} feasible paths in $\mathcal{G}$. We can write the global consistency objective for $\mathcal{G}$ as $
	\mathcal{L}_{\mathcal{G}} =  \sum_{p \in \mathcal{P}} \mathcal{L}_{p}^{\textit{\tiny{perceptual}}}$, where $p$ represents a path and $\mathcal{P}$ is the set of all feasible paths in $\mathcal{G}$.
	
	Optimizing the objective $\mathcal{L}_{\mathcal{G}}$ directly is intractable as it would require simultaneous training of all networks in $\mathcal{F}$ with a massive number of consistency paths\footnote{For example, a complete $\mathcal{G}$ with $n$ nodes includes $n(n-1)$ networks and $\sum_{k=2}^{L}{\binom{n}{k+1}}(k+1)!$ feasible paths, with path length capped at $L$.}.  
	In Alg.\ref{algorithm} we devise a straightforward training schedule for an approximate optimization of $\mathcal{L}_{\mathcal{G}}$.
	This problem is similar to inference in graphical models, where one is interested in marginal distribution of unobserved nodes given some observed nodes by passing ``messages'' between them through the graph until convergence. As exact inference is usually intractable for unconstrained graphs, often an approximate message passing algorithm with various heuristics is used.
	
	\vspace{-10pt}
	\begin{algorithm}
		\SetAlCapNameFnt{\scriptsize}
		\SetAlCapFnt{\scriptsize}
		\scriptsize
		\SetAlgoLined
		\KwResult{Trained edges $\mathcal{F}$ of graph $\mathcal{G}$}
		Train each $f {\in} \mathcal{F}$ independently. \Comment{initialization by standard direct training.} \\ \label{algo:initline} 
		\For{$k \gets 2$ to $L$}{        
			\While{\textit{LossConvergence}$(\mathcal{F})$ \textup{not met}}{    	            
				$f_{ij}{\leftarrow}$\textit{SelectNetwork}$(\mathcal{F})$ \Comment{selects target network to be trained.} \\ 
				$p{\leftarrow}$\textit{SelectPath}$(f_{ij},k,\mathcal{P})$ \Comment{selects a feasible consistency path for $f_{ij}$ with maximum length $k$ from $\mathcal{P}$.}\\         
				optimize $\mathcal{L}_{ijp}^{\textit{\tiny{perceptual}}}$ \Comment{trains$f_{ij}$ using path constraint $p$ in loss~\ref{eq:longpathloss}.}\\    	
		}}
		
		\caption{Globally Cross-Task Consistent Learning of Networks $\mathcal{F}$}
		\label{algorithm}
	\end{algorithm}
	\vspace{-10pt}
	Instead of optimizing all terms in $\mathcal{L}_{\mathcal{G}}$, Alg.\ref{algorithm} selects one network $f_{ij}{\in}\mathcal{F}$ to be trained, selects consistency path(s) $p{\in}\mathcal{P}$ for it, and trains $f_{ij}$ with $p$ for a fixed number of steps using loss~\ref{eq:longpathloss} (or its multi path version if multiple paths selected). This is repeated until all networks in $\mathcal{F}$ satisfy a convergence criterion. 
	
	\begin{figure*}
		\centering
		\includegraphics[trim={0mm 209mm 0mm 0},clip,width=1\textwidth]{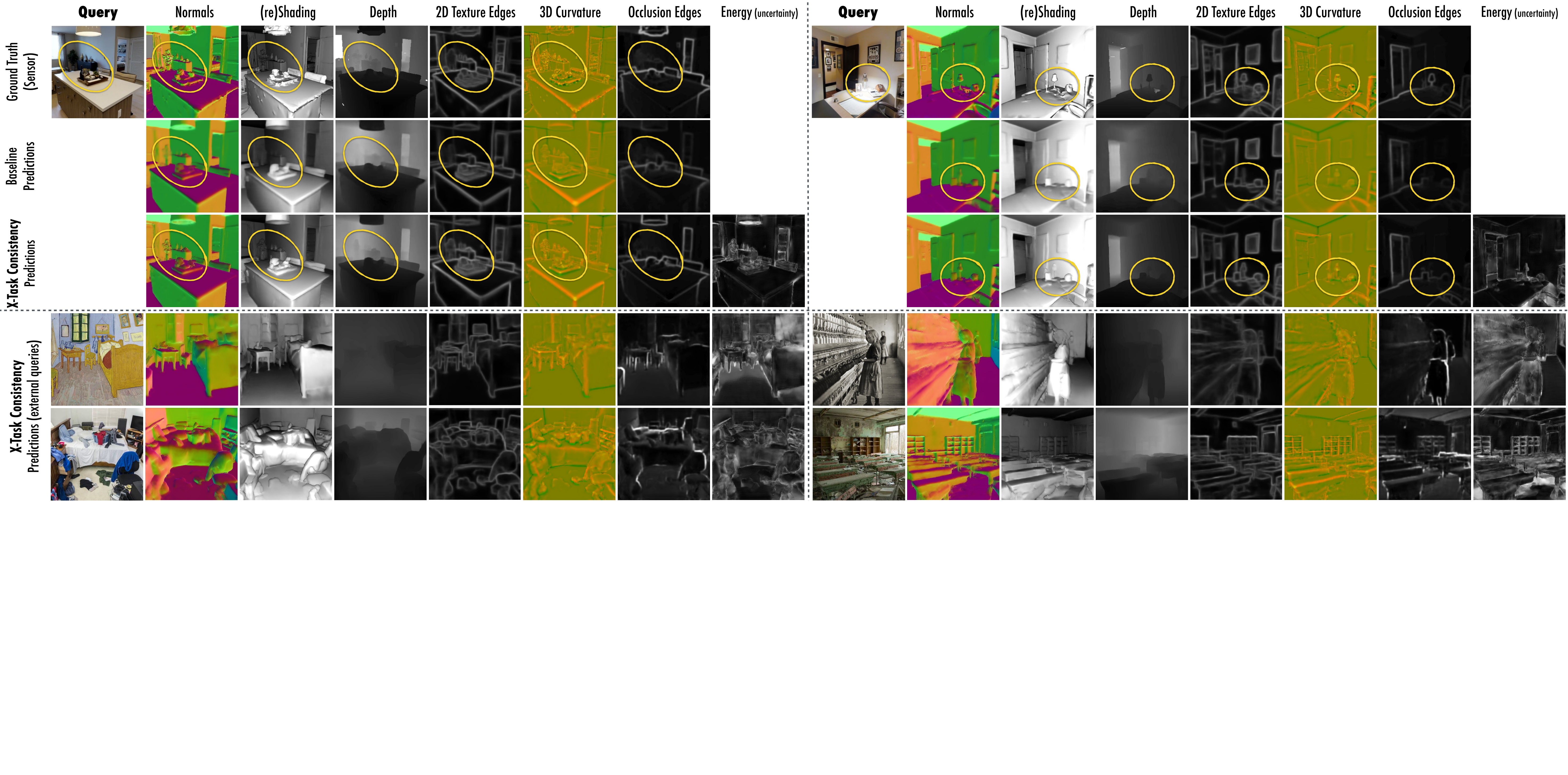}
		\vspace{-6.5mm}
		\caption{\footnotesize{\textbf{Qualitative results of predicting multiple domains along with the pixel-wise Consistency Energy.} The \textbf{top} queries are from the Taskonomy dataset's test set. The results of networks trained with consistency are more accurate, especially in fine-grained regions (zoom into the yellow markers), and more correlated across different tasks. The \textbf{bottom} images are external queries (no ground truth available) demonstrating the generalization and robustness of consistency networks to external data. Comparing the energy against a prediction domain (e.g., normals) shows that energy often correlates with error. More examples are provided on the \href{https://consistency.epfl.ch/}{project page}, and a live demo for user uploaded images is available at the \href{https://consistency.epfl.ch/demo}{demo page}. \textbf{External Queries}: \href{https://www.vangoghmuseum.nl/en/collection/s0047V1962?v=1}{Bedroom in Arles, Van Gogh} (1888); \href{http://100photos.time.com/photos/lewis-hine-cotton-mill-worker}{Cotton Mill Girl, Lewis Hine} (1908); \href{https://web.archive.org/web/20120428011323/http://www.boredpanda.com/chernobyl-20-years-after-the-accident/}{Chernobyl Pripyat Abandoned School} (c. 2009). [best seen on screen]}}
		\vspace{-1.5mm}
		\label{fig:qualitative_accuracy}
	\end{figure*}

	A number of choices for the selection criterion in \emph{SelectNetwork} and \emph{SelectPath} is possible, including round-robin and random selection. While we did not observe a significant difference in the final results, we achieved the best results using \emph{maximal violation} criterion: at each step select the network and path with the largest loss\footnote{\label{seesupmat}See \href{http://consistency.epfl.ch/supplementary_material}{supplementary material} for an experimental comparison.}. Also, Alg.\ref{algorithm} starts from shorter paths and progressively opens up to longer ones (up to length $L$) only after shorter paths have converged. This is based on the observation that the benefit of short and long paths in terms of enforcing cross-task consistency overlap, while shorter paths are computationally cheaper\textsuperscript{\ref{seesupmat}}. For the same reason, all of the networks are initialized by training using the standard direct loss (Op.\ref{algo:initline} in Alg.\ref{algorithm}) before progressively adding consistency terms. 
	
	Finally, Alg.\ref{algorithm} does not distinguish between $\mathcal{F}_x$ and $\mathcal{F}_x$ and can be used to train them all in the same pool. This means the selected path $p$ may include networks not fully converged yet. This is not an issue in practice, because, first, all networks are pre-trained with their direct loss (Op.\ref{algo:initline} in Alg.\ref{algorithm}) thus they are not wildly far from their convergence point. Second, the perceptual loss formulation makes training $f_{ij}$ robust to imperfections in functions in $p$ (Sec.~\ref{sec:perceploss}). 
	However, as practical applications primarily care about $\mathcal{F}_x$, rather than $\mathcal{F}_y$, one can first train $\mathcal{F}_y$ to convergence using Alg.\ref{algorithm}, then start the training of $\mathcal{F}_x$ with well trained and converged networks $\mathcal{F}_y$. We do the latter in our experiments.\footnote{A further cheaper alternative is applying cross-task consistent learning only on $\mathcal{F}_x$ and training $\mathcal{F}_y$ using standard independent training. This is significantly cheaper and more convenient, but still improves $\mathcal{F}_x$ notably.}
	Please see \href{http://consistency.epfl.ch/supplementary_material}{supplementary material} for how to normalize and balance the direct and consistency loss terms, as they belong to different domains with distinct numerical properties.

	\spacesaversection{{Consistency Energy}}
	\label{sec:energy_method}
	We quantify the amount of cross-task consistency in the system using an energy-based quantity~\cite{lecun2006tutorial} called \emph{Consistency Energy}. For a single query $x$ and domain $\ssmally_k$, the consistency energy is defined to be the standardized average of pairwise inconsistencies: 
	\begin{equation}
	\text{Energy}_{\ssmally_{k}}\hspace{-2pt}{(x)} ~\defeq~
	\hspace{-0.5pt}
	\raisebox{-1pt}{$
		\frac{ {1} }{ {|\ssmally|{{-}1}} }
		$}
	\hspace{-6pt}
	\raisebox{0pt}{$\displaystyle
		\sum_{\substack{\ssmally_i \in \ssmally, i\neq k}}
		$}
	\hspace{-9pt}
	\raisebox{-1pt}{$
		{
			\frac{ {|f_{\ssmally_i\ssmally_k}{\circ}f_{\ssmallx\ssmally_i}(x){-}f_{\ssmallx\ssmally_k}(x)| {-} \mu_{i}} }{ {\sigma_{i}} }
		}
		$},
	\label{eq:energy}
	\end{equation}
	where $\mu_{i}$ and $\sigma_{i}$ are the average and standard deviation of $|f_{\ssmally_i\ssmally_k}{\circ}f_{\ssmallx\ssmally_i}(x){-} f_{\ssmallx\ssmally_k}(x)|$ over the dataset. Eq.~\ref{eq:energy} can be computed per-pixel or per-image by average over its pixels. Intuitively, the energy can be thought of as the amount of variance in predictions in the lower row of Fig.~\ref{fig:qualitative_consistency} -- the higher the variance, the higher the inconsistency, and the higher the energy. 
	The consistency energy is an intrinsic quantity of the system and needs no ground truth or supervision. 
	
	In Sec.~\ref{sec:energy_results}, we show this quantity turns out to be quite informative as it can indicate the reliability of predictions (useful as a confidence/uncertainty metric) or a shift in the input domain (useful for domain adaptation). This is based on the fact that if the query is from the same data distribution as the training and is unchallenging, all inference paths of a system trained with consistency path constraints work well and yield similar results (as they were trained to); whereas under a distribution shift or for a challenging query, different paths break in different ways resulting in dissimilar predictions, and therefore, creating a higher variance. In other words, usually \emph{correct} predictions are \emph{consistent} while \emph{mistakes} are \emph{inconsistent}.  (Plots~\ref{fig:correlation},~\ref{fig:id_ood_energy_dist},~\ref{fig:error_energy_blur}.)

	\begin{figure*}
		\centering
		\includegraphics[trim={0mm 18mm 0mm 0},clip,width=1\textwidth]{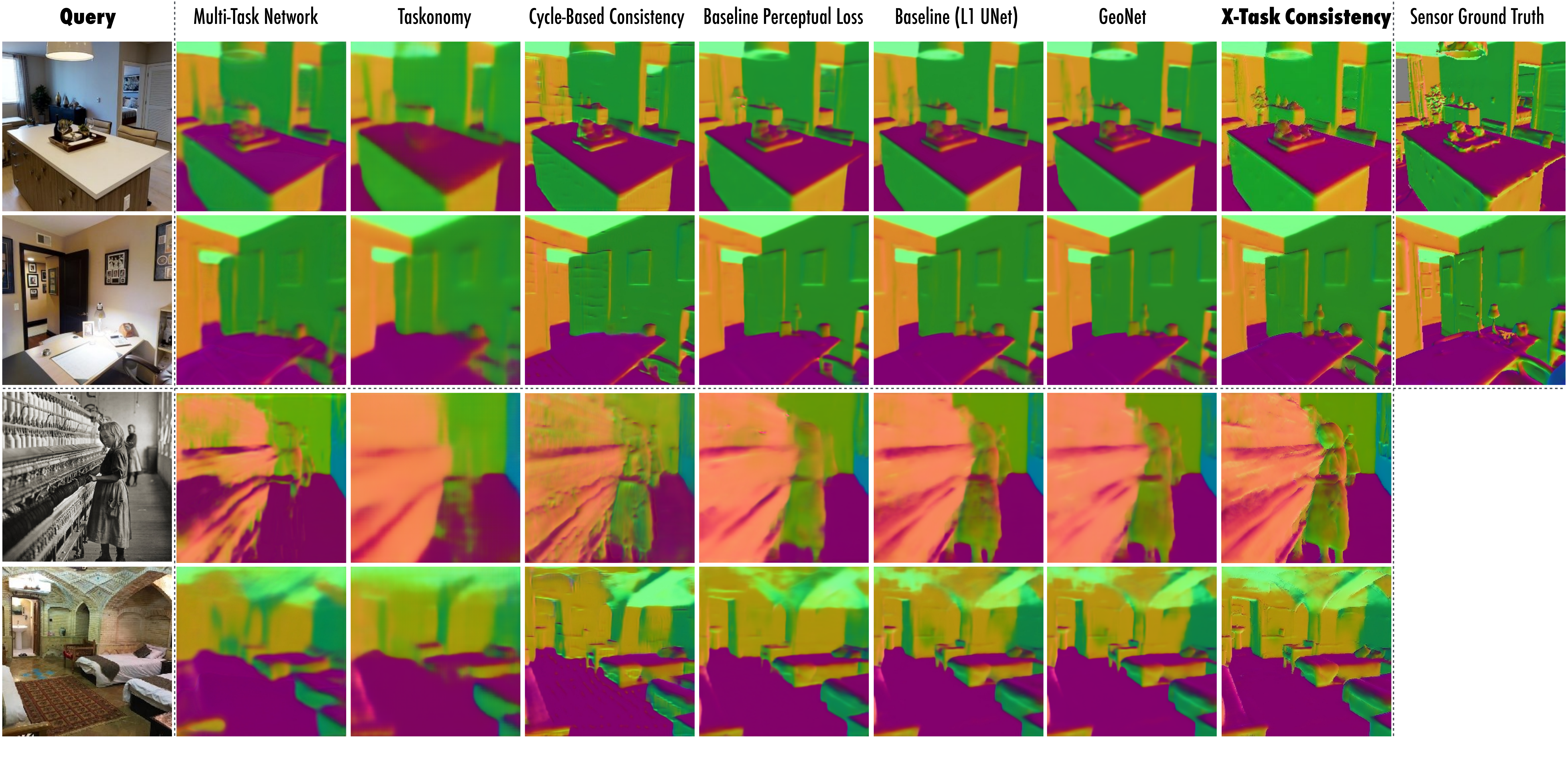} 
		\vspace{-6mm}		
		\caption{\footnotesize{\textbf{Learning with cross-task consistency vs various baselines} compared over surface normals. Queries are from Taskonomy dataset (\textbf{top}) or external data (\textbf{bottom}). Similar comparison for other domains and more images are provided on the \href{https://consistency.epfl.ch/}{project page}, and a live demo for user uploaded images is available at the \href{https://consistency.epfl.ch/demo}{demo page}. [best seen on screen]}}
		\vspace{-3mm}
		\label{fig:normalqualitative_accuracy}
	\end{figure*}

	\spacesaversection{Experiments}
	\label{sec:results}
	The evaluations are organized to demonstrate the proposed approach yields predictions that are \textbf{I.}  more \emph{consistent} (Sec.\ref{sec:results_consistency}), \textbf{II.} more \emph{accurate} (Sec.\ref{sec:results_accuracy}), and \textbf{III.} more \emph{generalizable} to out-of-training-distribution data (Sec.\ref{sec:results_generalization}). We also \textbf{IV.} quantitatively analyze the \emph{Consistency Energy} and report its utilities (Sec.\ref{sec:energy_results}).

	\vspace{2pt} \noindent	\textbf{Datasets:}
	We used the following datasets in the evaluations: 
	\vspace{-5pt}
	\begin{description}[leftmargin=2mm]
		\setlength\itemsep{-0.2em}
		\small{
			\item \textbf{Taskonomy~\cite{taskonomy18}:} We adopted Taskonomy as our main training dataset. It includes 4 million real images of indoor scenes with multi-task annotations for each image. The experiments were performed using the following 10 domains from the dataset: \emph{RGB images, surface normals, principal curvature, depth (zbuffer), reshading, {3D (occlusion) edges}, {2D (Sobel) texture edges}, {3D keypoints}, {2D keypoints}, and {semantic segmentation}}.  The tasks were selected to cover 2D, 3D, and semantic domains and have sensor-based/semantic ground truth. We report results on the test set. Also, as one of the out-of-domain tests, we use a version of Taskonomy images where they undergo distortions (e.g., blurring). 
			
			\item \textbf{Replica\cite{straub2019replica}} has high resolution 3D ground truth and enables more reliable evaluations of fine-grained details. We \emph{test} on 1227 images from Replica (no training), besides Taskonomy test data.

			\item \textbf{CocoDoom}~\cite{researchdoom}
			contains synthetic images from the \emph{Doom} video game. We use it as one of the out-of-training-distribution datasets.
			
			\item \textbf{ApolloScape}~\cite{appoloscape} contains real images of outdoor driving scenes. We use it as another out-of-training-distribution dataset.
			\item \textbf{NYU~\cite{Silberman2012}:} We also evaluated on NYUv2. The findings are similar to those on Taskonomy and Replica (in \href{http://consistency.epfl.ch/supplementary_material}{supplementary material}).		
		}
	\end{description}

	\vspace{0pt} \noindent	\textbf{Architecture \& Training Details:}
	We used a UNet~\cite{RonnebergerFB15} backbone architecture. We benchmarked alternatives, e.g., ResNet~\cite{resnet}, and found UNets to yield superior pixel-wise predictions. All networks in $\mathcal{F}_\ssmallx$ and $\mathcal{F}_\ssmally$ have a similar architecture. The networks have 6 down and 6 up sampling blocks and were trained using AMSGrad~\cite{reddi2019convergence} and Group Norm~\cite{wu2018group} with learning rate $3{\times}10^{-5}$, weight decay $2{\times}10^{-6}$, and batch size 32. Input and output images were linearly scaled to the range $[0,1]$ and resized down to $256\times256$. We used $\ell_1$ as the norm in all losses and set the max path length $L\text{=}3$. We experimented with different loss normalization methods and achieved the best results when the loss terms are weighted negative proportional to their respective gradient magnitude (details in \href{http://consistency.epfl.ch/supplementary_material}{supplementary material}).

	\begin{table*}
		\tiny
		\hspace{-0.005\textwidth}
		\begin{minipage}{1.025\textwidth}
			\input{tables/joint_tables}

		\end{minipage}
		\vspace{-6pt}	
		\captionof{table}{\footnotesize{\textbf{Quantitative Evaluation of Cross-Task Consistent Learning vs Baselines.} Results are reported on Replica and Taskonomy Datasets for four prediction tasks (normals, depth, reshading, pixel-wise semantic labeling) using `Direct' and `Perceptual' error metrics. The Perceptual metrics evaluate the target prediction in another domain (e.g., the \emph{leftmost} column evaluates the \emph{depth} inferred out of the predicted \emph{normals}). \textbf{Bold} marks the best-performing method. If more than one value is bold, their performances were statistically indistinguishable from the best, according to 2-sample paired t-test $\alpha=0.01$. Learning with consistency led to improvements with large margins in most columns. (In all tables, $\ell$ norm values are multiplied by 100 for readability. Methods that cannot be run for a given target are denoted by `$\times$'.})}\vspace{-2mm}
		\label{table_joint}
	\end{table*}
	
	\vspace{2pt} \noindent	\textbf{Baselines:}
	The main baseline categories are described below. To prevent confounding factors, our method and all baselines were implemented using the \emph{same UNet network} when feasible and were  \emph{re-trained on Taskonomy dataset}.
	\vspace{-4pt}
	\begin{description}[leftmargin=2mm]
		\setlength\itemsep{-0.2em}
		\small{
			\item \textbf{Baseline UNet (standard independent learning)} is the main baseline. It is identical to consistency models in all senses, except being trained with only the direct loss and no consistency terms.			
			\item \textbf{Multi-task learning:} A network with one shared encoder and multiple decoders each dedicated to a task, similar to~\cite{Kokkinos16}. This baseline shows if consistency across tasks would emerge by sharing a representation without explicit consistency constraints.
			\item  \textbf{Cycle-based consistency}, e.g.\cite{cycleGan17}, is a way of enforcing consistency between two domains \emph{assuming a bijection} between them. This assumption is violated between many domains (e.g. RGB$\leftrightarrow$3D, as texture cannot be recovered from 3D). This baseline is a special case of the triangle in Fig.\ref{fig:graph}(b) by setting $\smally_2{=}\smallx$. 
			\item  \textbf{Baseline perceptual loss network} uses frozen random (Gaussian weight) networks as $\mathcal{F}_\ssmally$, rather than training them to be cross-task functions. This baseline would show if the improvements were owed to the priors in the architecture of constraint networks, rather than them executing cross-task consistency constraints.
			\item  \textbf{GAN-based image translation:} We used Pix2Pix~\cite{pix2pix}, which is conditional GAN based framework~\cite{mirza2014conditional}.
			\item  \textbf{Blind guess:}\hspace{-2pt} A query-agnostic statistically informed guess computed from data for each domain (visuals in \href{http://consistency.epfl.ch/supplementary_material}{supplementary}). It shows what can be learned from general dataset regularities.~\cite{taskonomy18}
			\item  \textbf{GeoNet~\cite{Geonet18}} is a task-specific consistency method analytically curated for depth and normals. This baseline shows how closely the task-specific consistency methods based on known analytical relationships perform vs the proposed generic data-driven method. The ``original'' and ``updated'' variants represent original authors' released networks and our re-implemented and re-trained version. 
		}
	\end{description}

	\subsection{Consistency of Predictions}
	\label{sec:results_consistency}
	Fig.~\ref{fig:energy_over_time} (blue) shows the amount of inconsistency in test set predictions (Consistency Energy) successfully decreases over the course of training. The convergence point of the network trained with consistency constraints is well below baseline independent learning (orange) and multi-task learning (green)--which shows consistency among predictions \emph{does not naturally emerge} in either case without explicit constraining. Plots of individual loss terms similarly show minimizing the direct term does not lead to automatic minimization of consistency terms (provided in \href{http://consistency.epfl.ch/supplementary_material}{supplementary}).

	\subsection{Accuracy of Predictions}
	\label{sec:results_accuracy}	
	Figures~\ref{fig:qualitative_accuracy} and ~\ref{fig:normalqualitative_accuracy} compare the prediction results of networks trained with cross-task consistency against the baselines in different domains. The improvements are considerable particularly around the difficult \emph{fine-grained details}. 
	
	Quantitative evaluations are provided in Tab.~\ref{table_joint} for Replica dataset and Taskonomy datasets on depth, normal, reshading, and pixel-wise semantic prediction tasks. 
	Learning with consistency led to large improvements in most of the setups.
	As most of the pixels in an image belong to easy to predict regions governed by the room layout (e.g., ceiling, walls), the standard pixel-wise error metrics (e.g., $\ell_1$) are dominated by them and consequently insensitive to fine-grained changes. Thus, besides standard \emph{Direct} metrics, we report \emph{Perceptual} error metric (e.g., \emph{normal$\shortrightarrow$curvature}) that evaluate the same prediction, but with a non-uniform attention to pixel properties.\footnote{For example, evaluation of normals via the \emph{normal$\shortrightarrow$curvature} metric is akin to paying more attention to where normals change, hence reducing the domination of flat regions, such as walls, in the numbers.}
	Each perceptual error provides a different angle, and the optimal results would have a low error for \emph{all} metrics.

	The corresponding \emph{Standard Error} for the reported numbers are provided in \href{http://consistency.epfl.ch/supplementary_material}{supplementary material}, which show the trends are statistically significant. 
	Tab.~\ref{table_joint} also includes evaluation of the networks when trained with little data (0.25\% subset of Taskonomy dataset), which shows the consistency constraints are useful under low-data regime as well. 	
	
	We adopted normals as the canonical task for more extensive evaluations, due to its practical value and abundance of baselines. The conclusions remained the same regardless. 		
	
	\textbf{Using Consistency with Unsupervised Tasks}: 
	Unsupervised tasks can provide consistency constraints, too. Examples of such tasks are 2D Edges and 2D Keypoints (SURF\cite{bay2006surf}), which are included in our dictionary. Such tasks have fixed operators that can be applied on any image to produce their respective domains without any additional supervision. Interestingly, we found enforcing consistency with these domains is still useful for gaining better results (see \href{http://consistency.epfl.ch/supplementary_material}{supplementary material} for the experiment). The ability to utilize unsupervised tasks further extends the applicability of our method to single/few task datasets.

	\subsection{Utilities of {{Consistency Energy}}}
	\label{sec:energy_results}
	Below we quantitatively analyze the Consistency Energy. The energy is shown (per-pixel) for sample queries in Fig.~\ref{fig:qualitative_accuracy}.

\begin{figure}
		\centering
		\subfigure[{Energy During Training}]{
			\includegraphics[width=0.205\textwidth]{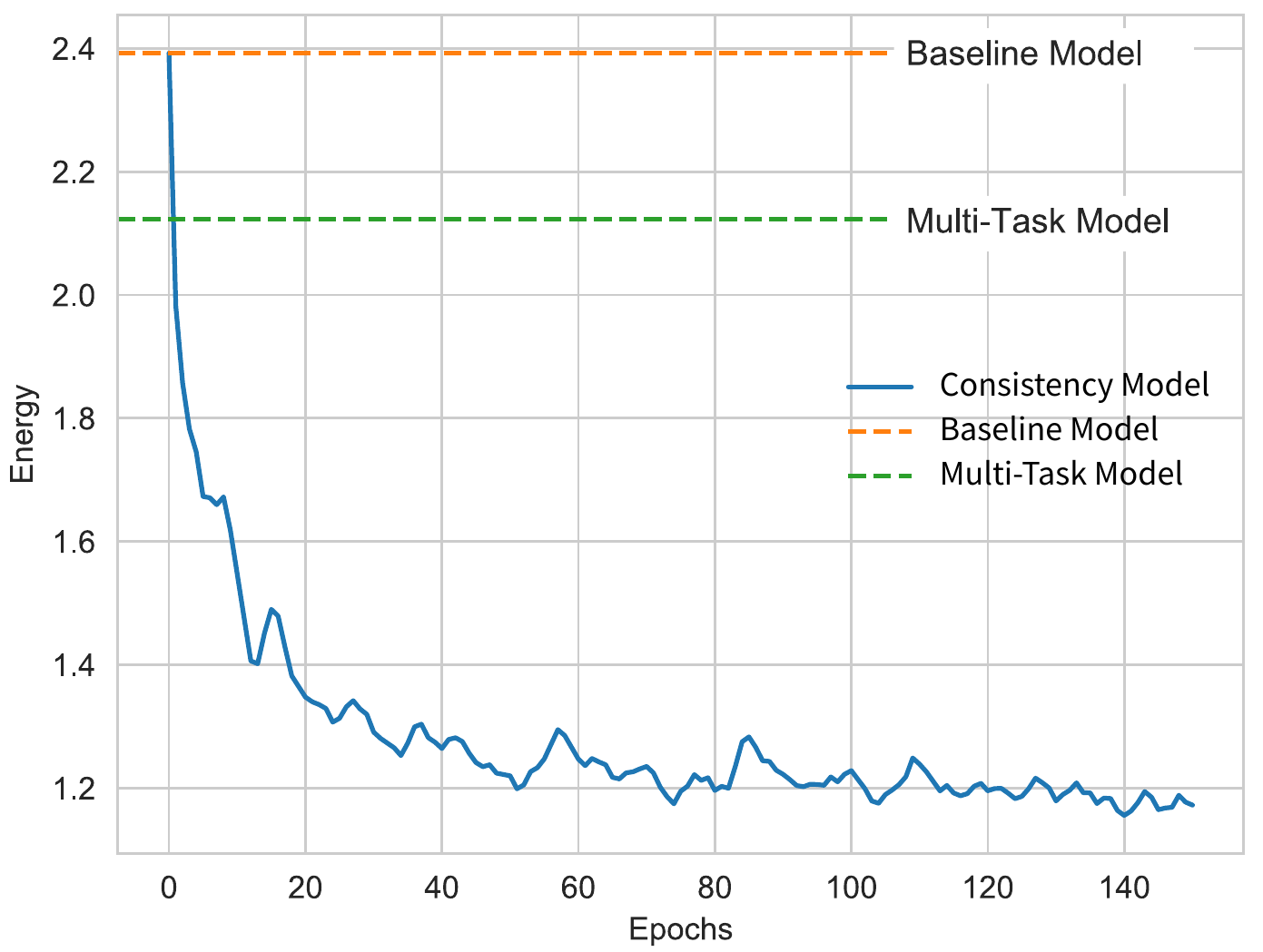}
			\label{fig:energy_over_time}
		}		
		\subfigure[{Energy\scriptsize{ vs }\footnotesize{Error}}]{
			\includegraphics[width=0.23\textwidth]{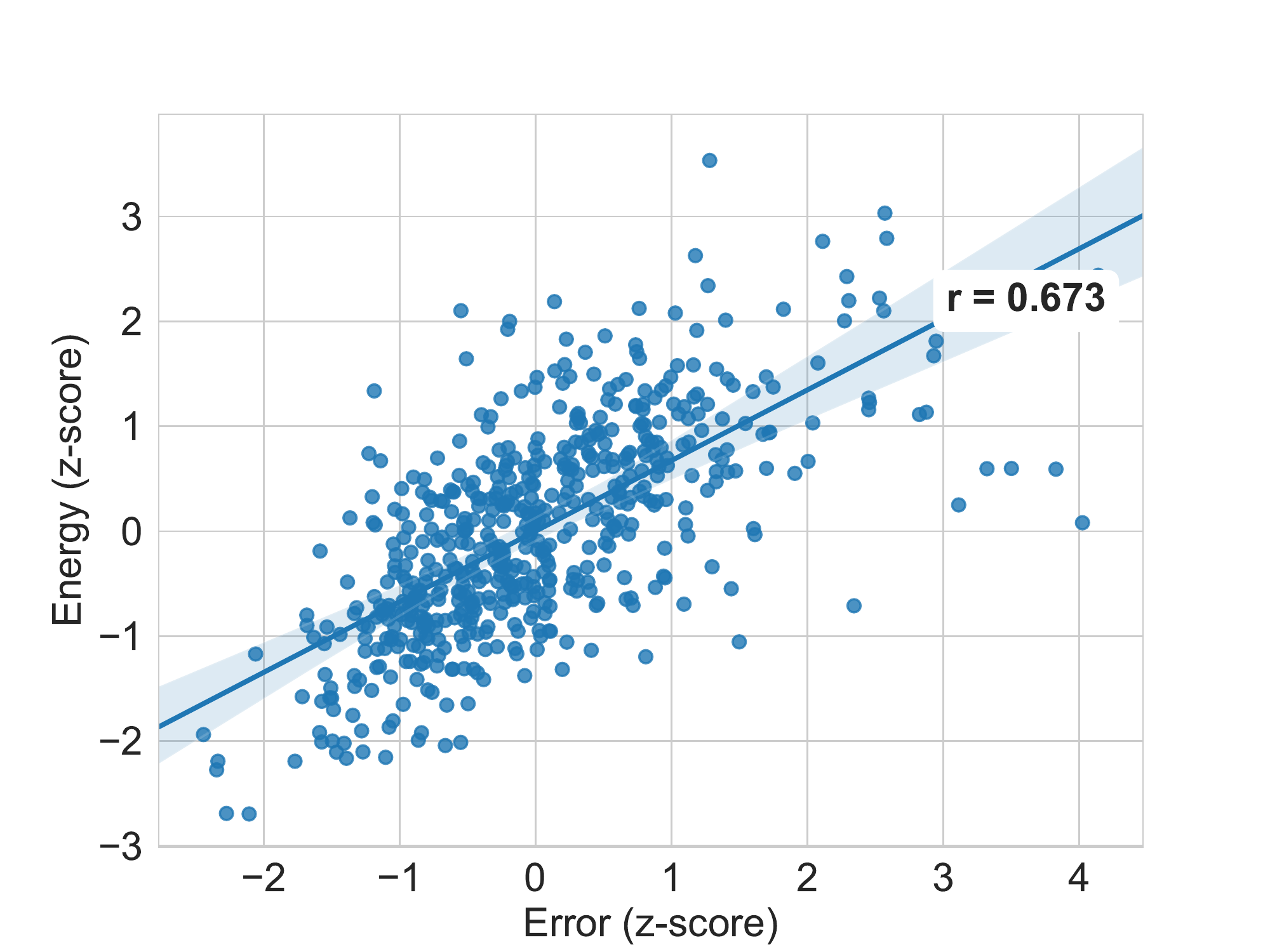}
			\label{fig:correlation}
		}
		\subfigure[{\hspace{-3.2pt}\footnotesize{Energy}\scriptsize{ vs }\footnotesize{Discrete Domain Shift}}]{
			\includegraphics[width=0.21\textwidth]{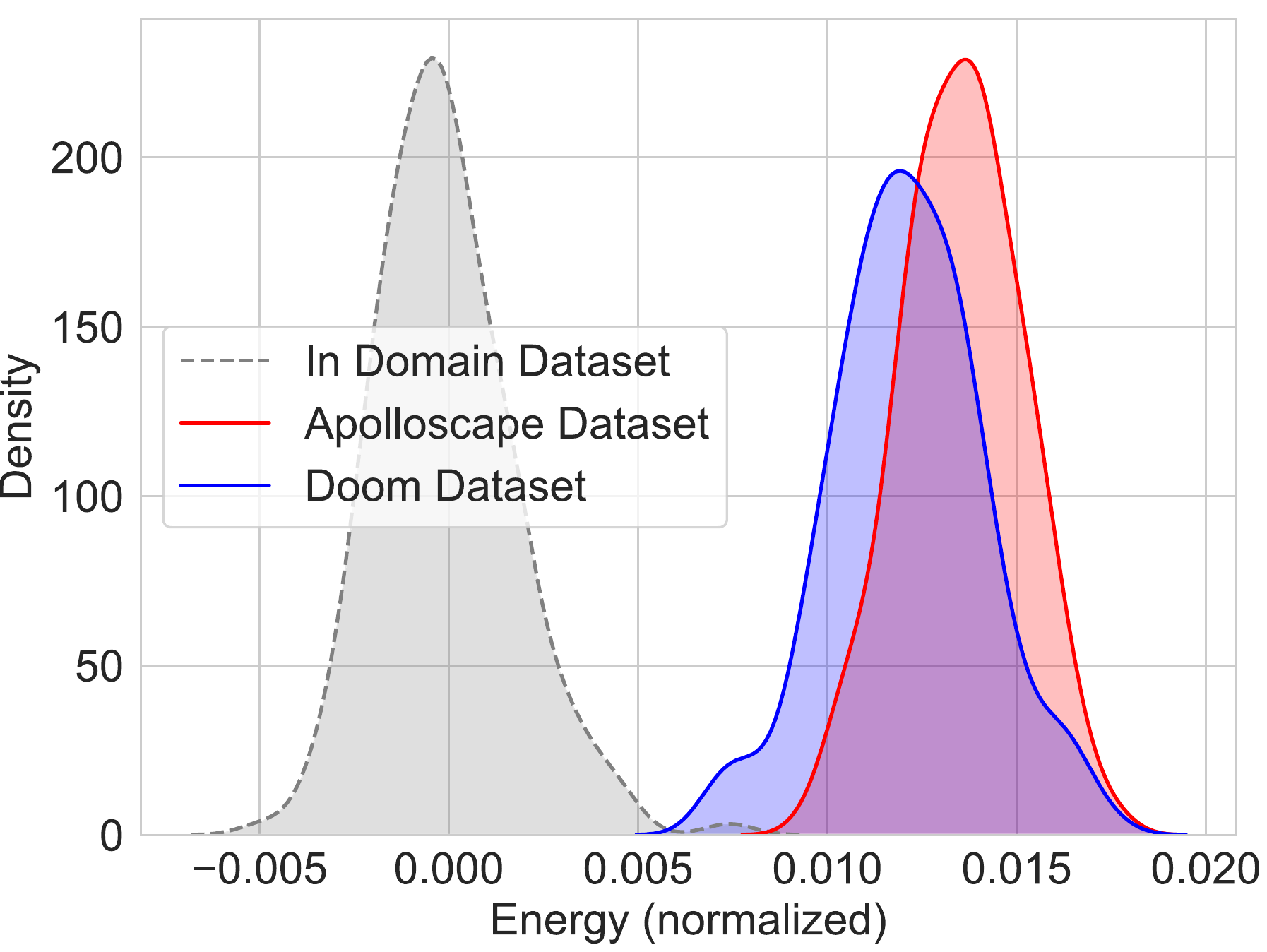}
			\label{fig:id_ood_energy_dist}
		} 
		\subfigure[\hspace{-2pt}{Energy\scriptsize{ vs }\footnotesize{Continuous Domain Shift}}]{
			\includegraphics[width=0.24\textwidth]{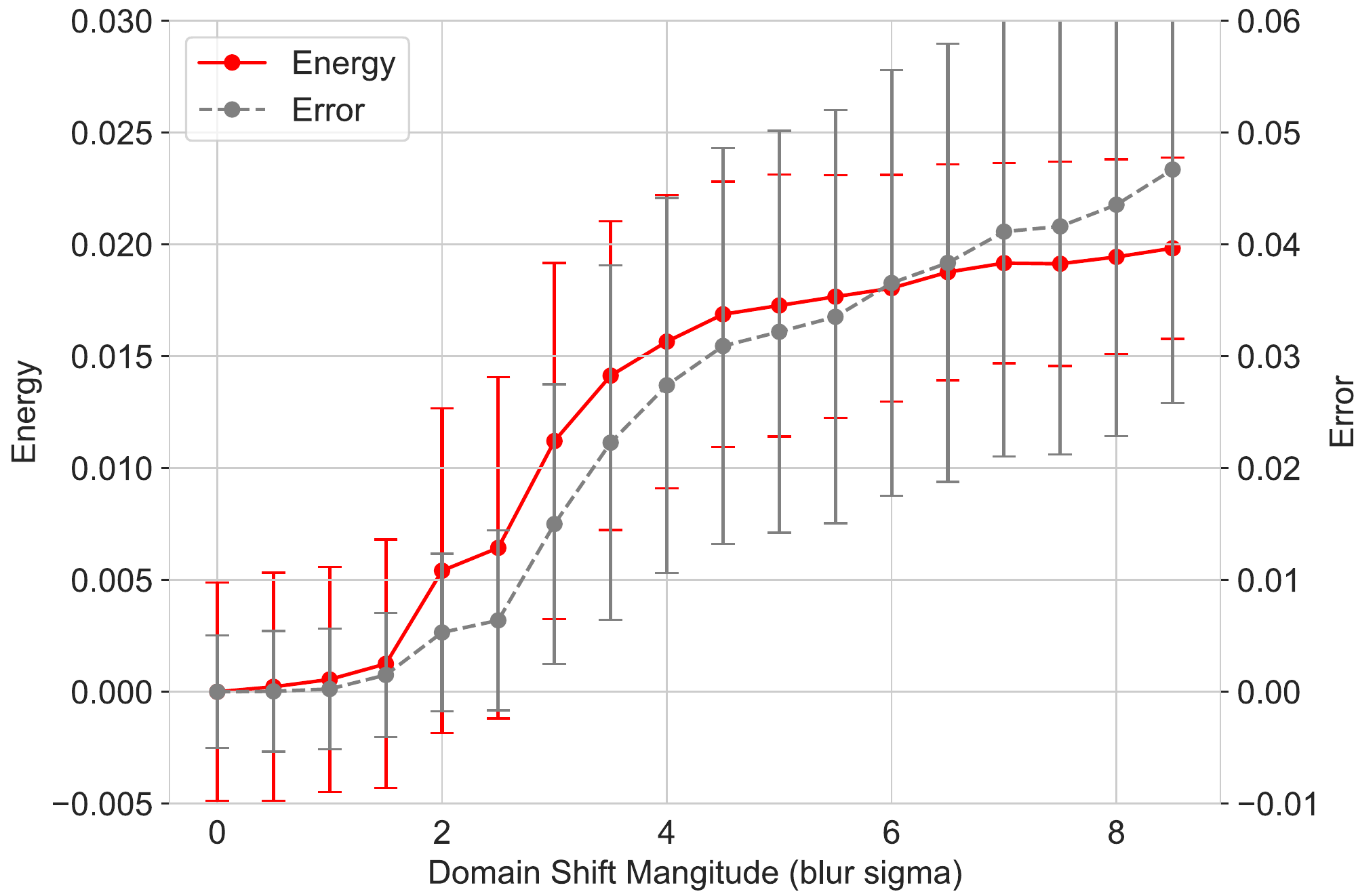}
			\label{fig:error_energy_blur}
		} \vspace{-14pt}
		\caption{\footnotesize{\textbf{Analyses of Consistency Energy.} }}
		\label{fig:energy}
	\end{figure}

	\textbf{Consistency Energy as a Confidence Metric (Energy vs Error):} Plot~\ref{fig:correlation} shows the energy of predictions has a strong positive correlation with the error computed using ground truth (Pearson corr. 0.67). This suggests the energy can be adopted for confidence quantification and handling uncertainty. This experiment was done on Taskonomy test set thus images had no domain shift from the training data.

	\begin{table*}
		\begin{minipage}[]{\textwidth}
			\begin{minipage}[]{0.4\textwidth}
				\centering
				\vspace{-1.9mm}
				\includegraphics[width=.56\columnwidth]{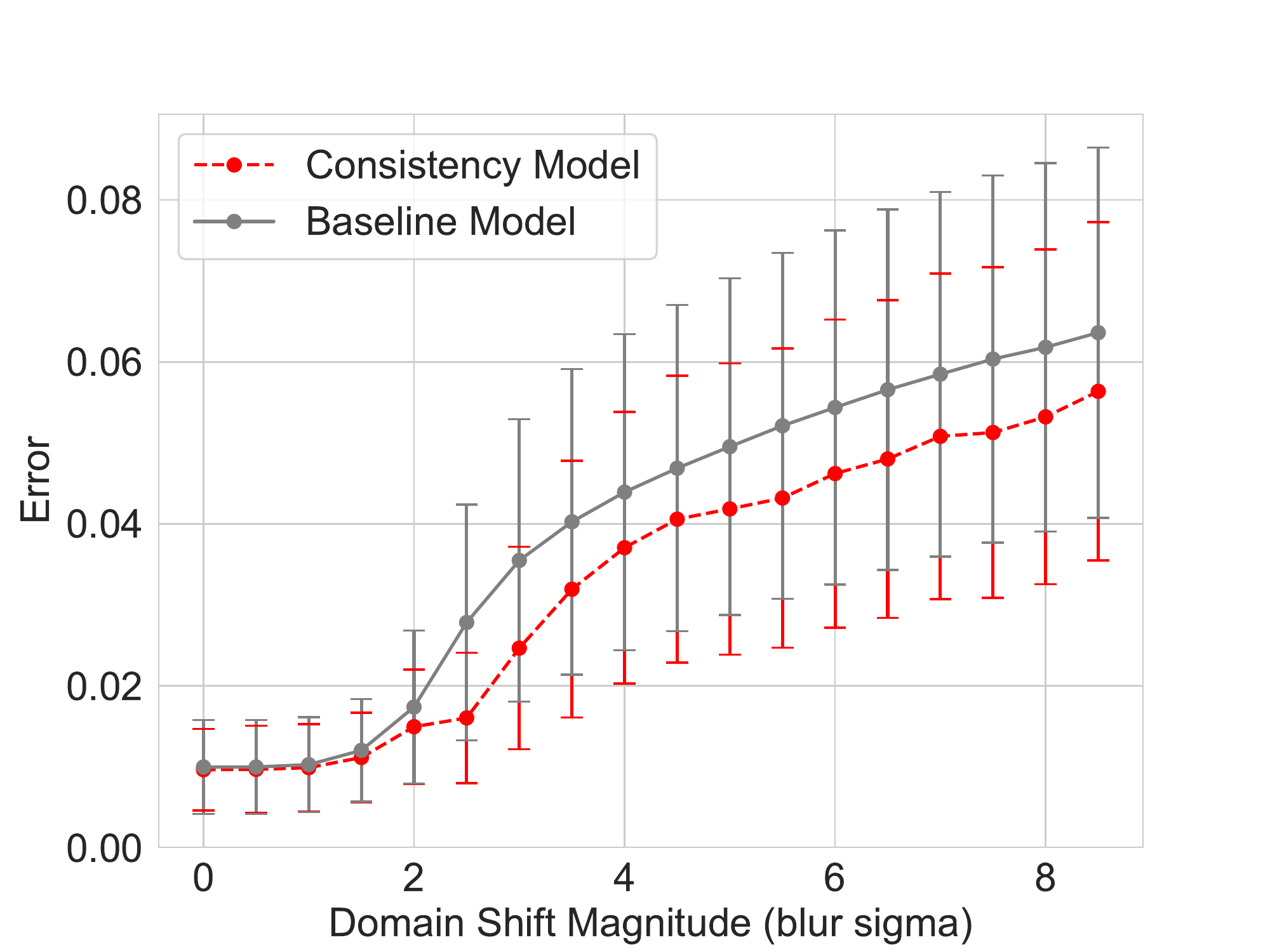}
				\vspace{-1.5mm}
				\captionof{figure}{\footnotesize{\textbf{Error with Increasing (Smooth) Domain Shift.} The network trained with consistency is more robust to the shift.}}\label{fig:ood_quantative}
			\end{minipage}
			\hfill
			\begin{minipage}[]{0.58\textwidth}
				\centering
				\scriptsize
				\addtolength{\tabcolsep}{.5pt}
				\begin{tabular}{|c|ccc|cc|}
					\hline
					\multicolumn{1}{|c|}{} & \multicolumn{3}{c|}{Error (Post-Adaption)} & \multicolumn{2}{c|}{Error (Pre-Adaptation)} \\
					\cline{2-6}
					Novel Domain & \# images & Consistency & Baseline & Consistency & Baseline \\
					\hline
					\multirow{2}{40pt}{\centering Gaussian blur \tiny{(Taskonomy)}} & 128 & 17.4 (+14.7\%) & 20.4  & \multirow{2}{*}{46.2 (+12.8\%)} & \multirow{2}{*}{53.0}  \\
					& 16 & 22.3 (+8.6\%) & 24.4  &  &  \\
					\hline
					\multirow{2}{40pt}{\centering CocoDoom} & 128 & 18.5 (+19.2\%) & 22.9  & \multirow{2}{*}{54.3 (+15.8\%)}& \multirow{2}{*}{64.5} \\
					& 16 & 27.1 (+24.5\%) & 35.9  &  &  \\
					\hline
					\centering ApolloScape & 8 & 40.5 (+11.9\%) & 46.0  & \multirow{1}{*}{55.8 (+5.5\%)} & \multirow{1}{*}{59.1}  \\
					\hline					
				\end{tabular}\vspace{.5mm}
				\captionof{table}{\footnotesize{\textbf{Domain generalization and adaptation} on CocoDoom, ApolloScape, and Taskonomy blur data. Networks trained with consistency show better generalization to new domains and a faster adaptation with little data. (relative improvement in parentheses)}} \label{table:ood_quantative}
			\end{minipage}
		\end{minipage}
	\end{table*}

	\textbf{Consistency Energy as a Domain Shift Detector:}
	Plot~\ref{fig:id_ood_energy_dist} shows the energy distribution of in-distribution (Taskonomy) and out-of-distribution datasets (ApolloScape, CocoDoom). Out-of-distribution datapoints have notably higher energy values, which suggests that energy can be used to detect anomalous samples or domain shifts. Using the per-image energy value to detect out-of-distribution images achieved $\text{ROC-AUC}\text{=}0.95$; the out-of-distribution detection method OC-NN~\cite{chalapathy2018anomaly} scored $0.51$.

	Plot~\ref{fig:error_energy_blur} shows the same concept as \ref{fig:id_ood_energy_dist} (energy vs domain shift), but when the shift away from the training data is smooth. The shift was done by applying a progressively stronger Gaussian blur with kernel size 6 on Taskonomy test images. The plot also shows the error computed using ground truth which has a pattern similar to the energy.

	We find the reported utilities noteworthy as handling uncertainty, domains shifts, and measuring prediction confidence in neutral networks are open topics of research~\cite{ovadia2019trust,guo2017calibration} with critical values in, e.g., active learning~\cite{sener2017active}, real-world decision making~\cite{Kochenderfer:2015:DMU:2815660}, and robotics~\cite{proctor2017tolerances}.

	\begin{figure}
		\centering
		\vspace{-3mm}
		\includegraphics[width=1\columnwidth]{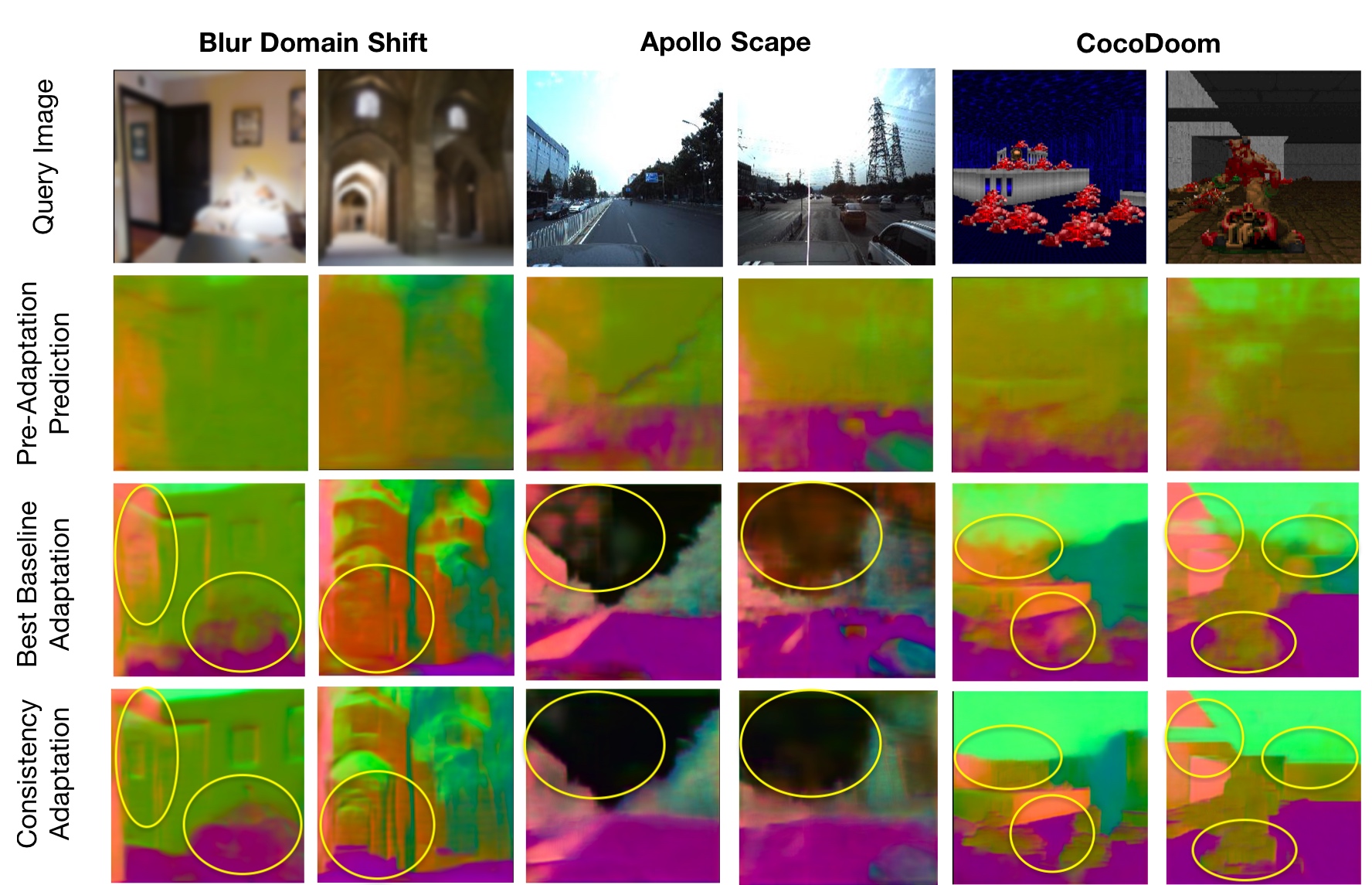}
		\vspace{-7mm}		
		\caption{\footnotesize{\textbf{Domain adaptation} results shown for three target domains (ApolloScape~\cite{appoloscape}, CocoDoom~\cite{researchdoom}, Gaussian-blur Taskonomy~\cite{taskonomy18}). Networks trained with consistency show better adaptation with little data. \vspace{-0pt}}}
		\vspace{-2mm}		
		\label{fig:ood_qualitative}
	\end{figure}

	\subsection{Generalization \& Adaptation to New Domains}
	\label{sec:results_generalization}	
	To study: \textbf{I.} how well the networks generalize to new domains without any adaptation and quantify their resilience, and \textbf{II.} how efficiently they can adapt to a new domain given a few training examples by fine-tuning, we test the networks trained on Taskonomy dataset on various new domains.
	The experiment were conducted on smooth (blurring~\cite{jo2017measuring}) and discrete (Doom~\cite{researchdoom}, ApolloScape~\cite{appoloscape}) shifts. For  (\textbf{II}), we use a small number (16-128) of images from the new domain to fine-tune the networks with and without consistency constraints. The original training data (Taskonomy) is retained during fine-tuning so prevent the networks from forgetting the original domain~\cite{learningWithoutForgetting}.

	Models trained with consistency constraints generally show more robustness against domain shifts (see Fig.~\ref{fig:ood_quantative} and pre-adaptation numbers in Table~\ref{table:ood_quantative}) and a better adaptation with little data (see post-adaptation numbers in Table~\ref{table:ood_quantative} and Fig.~\ref{fig:ood_qualitative}). The challenging external queries shown in Figures~\ref{fig:qualitative_accuracy}\&\ref{fig:normalqualitative_accuracy}\&\ref{fig:motiv} similarly denote a good generalization.

	\vspace{2pt} \noindent	\textbf{Supplementary Material:}
	We defer additional discussions and experiments, particularly analyzing different aspects of the optimization, stability analysis of the experimental trends, and proving qualitative results at scale to the \href{http://consistency.epfl.ch/supplementary_material}{supplementary material} and the \href{http://consistency.epfl.ch/}{project page}.

	\spacesaversection{Conclusion and Limitations}
	We presented a general and data-driven framework for augmenting standard supervised learning with cross-task consistency. The evaluations showed learning with cross-task consistency fits the data better yielding more accurate predictions and leads to models with improved generalization. The Consistency Energy was found to be an informative intrinsic quantity with utilities toward confidence estimation and domain shift detection. 
	Below we briefly discuss some of the limitations and assumptions:

	\textbf{Path Ensembles}: We used the various inference paths only as a way of enforcing consistency. Aggregation of \emph{multiple} (comparably weak) inference paths into a \emph{single strong} estimator (e.g., in a manner similar to boosting) is a promising direction that this paper did not address. Performing the aggregation in a probabilistic manner seems viable, as we found the errors of different paths are sufficiently uncorrelated, suggesting possibility of assembling a strong estimator. 
	
	\textbf{Unlabeled/Unpaired Data}: The current framework requires paired training data. Extending the concept to unlabeled/unpaired data, e.g., as in~\cite{cycleGan17}, appears feasible and remains open for future work.

	\textbf{Categorical/Low-Dimensional Tasks}: We primarily experimented with pixel-wise tasks. Classification tasks, and generally tasks with low-dimensional outputs, will be interesting to experiment with, especially given the more severely ill-posed cross-task relationships they induce.

	\textbf{Optimization Limits}: The improvements gained by incorporating consistency are bounded by the success of available optimization techniques, as addition of consistency constrains at times makes the optimization job harder. Also, implementing cross-task functions as neural networks makes them subject to certain \textbf{output artifacts} similar to those seen in image synthesis with neural networks.
	
	\textbf{Adversarial Robustness}: Lastly, if learning with cross-task consistency indeed reduces the tendency of neural networks to learn surface statistics~\cite{jo2017measuring} (Sec.~\ref{sec:intro}), studying its implications in defence against adversarial attacks will be worthwhile.
	
	\textbf{Energy Analyses}: We performed post-hoc analyses on the Consistency Energy. More concrete understanding of the properties of the energy and potentially using it actively for network modification, e.g, in unsupervised domain adaptation, requires further focused studies.

	\vspace{3pt}
	{\noindent\textbf{Acknowledgement:} This work was supported by a grant from SAIL Toyota Center for AI Research\footnote{Toyota Research Institute (``TRI'') provided funds to assist the authors with their research but this article solely reflects the opinions and conclusions of its authors and not TRI or any other Toyota entity.}, a Vannevar Bush Faculty Fellowship, ONR MURI  grant N00014-14-1-0671, an Amazon AWS Machine Learning Award, and Google Cloud.}
	
	
	{\small
		\bibliographystyle{ieee_fullname}
		\bibliography{egbib}
	}
	
\end{document}

%% file: spacesaver_bib.tex
\belowdisplayskip \abovedisplayskip

\usepackage{letltxmacro}

\LetLtxMacro{\oldsection}{\section}
\newcommand{\spacesaversection}[1]{
	\vspace{-0.08in}
	\oldsection{#1}
	\vspace{-0.06in}
}

\LetLtxMacro{\oldsubsection}{\subsection}
\renewcommand{\subsection}[1]{
    \vspace{-0.06in}
    \oldsubsection{#1}
    \vspace{-0.07in}
}

\LetLtxMacro{\oldsubsubsection}{\subsubsection}
\renewcommand{\subsubsection}[1]{
    \vspace{-0.10in}
    \oldsubsubsection{#1}
    \vspace{-0.08in}
}

%% file: tables/joint_tables.tex
\addtolength{\tabcolsep}{-5pt}\begin{tabular}{|c|cc|c|cc|c|cc|c|cccc|c|cccc|c|cccc|c|c|}
        \hline
        \multicolumn{1}{|c|}{\multirow{4}{*}{
        \backslashbox{~~Method}{\hspace{-0mm}\vspace{4.1mm}{~~~~~~~~~~Setup~~}}
        }} 
        & \multicolumn{9}{c|}{Replica Dataset} 
        & \multicolumn{16}{c|}{Taskonomy Dataset} \\
        \cline{2-26}
        & \multicolumn{3}{c|}{\textbf{Normals} } 
        & \multicolumn{3}{c|}{\textbf{Depth} } 
        & \multicolumn{3}{c|}{\textbf{reShading} } 
        & \multicolumn{5}{c|}{\textbf{Normals} } 
        & \multicolumn{5}{c|}{\textbf{Depth} } 
        & \multicolumn{5}{c|}{\textbf{reShading} } 
        & \multicolumn{1}{c|}{\textbf{Semantic Segm.}} \\
        & \multicolumn{2}{c|}{Perceptual Err.} & \multicolumn{1}{c|}{Direct}
        & \multicolumn{2}{c|}{Perceptual Err.} & \multicolumn{1}{c|}{Direct}
        & \multicolumn{2}{c|}{Perceptual Err.} & \multicolumn{1}{c|}{Direct}
        & \multicolumn{4}{c|}{Perceptual Err.} & \multicolumn{1}{c|}{Direct}
        & \multicolumn{4}{c|}{Perceptual Err.} & \multicolumn{1}{c|}{Direct}
        & \multicolumn{4}{c|}{Perceptual Err.} & \multicolumn{1}{c|}{Direct}
        & \multicolumn{1}{c|}{Direct} \\
        & Depth & reShade  & $\ell_1$ Err.
        & Norm. & reShade  & $\ell_1$ Err.
        & Norm. & Depth  & $\ell_1$ Err.
        & Depth & reShade & Curv. & \multicolumn{1}{c|}{Edge(2D)} & $\ell_1$ Err.
        & Norm. & reShade & Curv. & \multicolumn{1}{c|}{Edge(2D)} & $\ell_1$ Err.
        & Norm. & Depth & Curv. & \multicolumn{1}{c|}{Edge(2D)} & $\ell_1$ Err.
        & X-Entropy $(\downarrow)$ \\
        \hline
        \hline
        Blind Guess    &
                4.75   &         33.31  &         16.02  & 
                22.23  &         19.94  &         4.81   &
                15.74  &         5.14   &         16.45  & 
                7.39   &         38.11  &         3.91   &         12.05  &         17.77  &
                22.37  &         27.27  &         7.96   &         12.77  &         7.07   &
                19.96  &         7.14   &         3.53   &         12.62  &         24.85  &
                \multirow{3}{*}{} \\
        Taskonomy Networks     &
                3.73   &         11.07  &         6.55   & 
                18.06  &         15.39  &         3.72   & 
                8.70   &         3.85   &         11.43  & 
                7.19   &         22.68  &         3.68   &         10.70  &         7.54   & 
                18.82  &         20.83  &         6.65   &         14.10  &         4.55   &
                11.72  &         4.69   &         3.54   &         11.19  &         16.58  &
                 \\
        Multi-Task      &
                5.58   &         22.11  &         6.03   & 
                15.30  &         16.14  &         2.44   & 
                7.24   &         3.36   &         10.32  & 
                8.78   &         27.32  &         3.65   &         10.16  &         7.07   &
                17.18  &         19.55  &         7.54   &         13.67  &         2.81   &
                9.19   &         3.54   &         3.56   &         10.75  &         11.61  &
                  \\
            \cline{26-26}
        GeoNet  (original)                 &
                6.23   &         19.34  &          7.48  & 
                13.88  &         14.03  &          4.01  & 
                \na    &         \na    &          \na   &
                7.71   &         27.35  &  \textbf{3.32} &  \textbf{9.09} &        9.58   & 
                15.44  &         18.73  &          4.03  &         10.78  &        4.07   & 
                \na    &         \na    &          \na   &         \na    &        \na    &
                \na    \\
         \cline{5-10} \cline{16-26}
         Cycle Consistency         &
                5.65   &         22.39  &         7.13   & 
                \multicolumn{6}{c|}{\multirow{3}{*}{{}}} & 
                8.81   &         30.33  &         3.84   &         10.26  &         8.68   & 
                \multicolumn{11}{c|}{\multirow{3}{*}{}}\\
        Baseline Perceptual Loss            &
                4.88   &         15.34  &         4.99   & 
                \multicolumn{6}{c|}{} & 
                8.59   &         23.98  &         3.41   &         10.01  &         6.17   & 
                \multicolumn{11}{c|}{} \\
        Pix2Pix                         &
                4.52   &         19.03  &         7.70   & 
                \multicolumn{6}{c|}{} & 
                8.12   &         26.23  &         3.83   &         10.33  &         9.40   & 
                \multicolumn{11}{c|}{} \\
        \cline{5-10}    \cline{16-26}
         Baseline UNet ($\ell_1$)                &
                4.69   &         13.15  &          4.96  & 
                10.47  &         12.99  &          1.99  & 
                6.90   &         2.74   &  \textbf{9.55} & 
                8.17   &         20.94  &          3.41  &         9.98   & \textbf{5.95}  &
                13.62  &         15.68  &          7.31  &         12.61  & \textbf{2.27}  &
                9.58   &         3.38   &          3.78  &         10.85  & \textbf{10.45} &
                0.246  \\
        GeoNet  (updated)                 &
                4.62   &         12.79  &  \textbf{4.70} & 
                10.47  &         12.75  &          1.83  & 
                \na    &         \na    &          \na   &
                8.18   &         20.84  &          3.40  &         9.99   & \textbf{5.91}  & 
                13.77  &         15.76  &          7.52  &         12.67  & \textbf{2.26}  & 
                \na    &         \na    &          \na   &         \na    &         \na    &
                \na    \\
        \textbf{X-Task Consistency}     &
                \textbf{2.07} & \textbf{9.99}  & \textbf{4.80} & 
                \textbf{7.01} & \textbf{11.21} & \textbf{1.63} & 
                \textbf{5.50} & \textbf{1.96}  & \textbf{9.22} & 
                \textbf{4.32} & \textbf{12.15} & \textbf{3.29} &         9.50  &         6.08   &
                \textbf{9.46} & \textbf{12.66} & \textbf{3.61} & \textbf{9.82} & \textbf{2.29}  &
                \textbf{7.13} & \textbf{2.51}  & \textbf{3.28} & \textbf{9.38} & \textbf{10.52} &
                \textbf{0.237}  \\
        \hline
                \hline
        0.25\% Data: Baseline           &
                5.65   &         21.76  &         7.61  & 
                \multicolumn{6}{c|}{\multirow{2}{*}{}} & 
                8.86   &         26.91  &         3.78  &         10.31  & \textbf{8.17}  &
                \multicolumn{11}{c|}{\multirow{2}{*}{}}  \\
        \textbf{0.25\% Data: Consistency}        &
                \textbf{2.41} & \textbf{12.26} & \textbf{7.28} & 
                \multicolumn{6}{c|}{} & 
                \textbf{5.07} & \textbf{15.96} & \textbf{3.74} & \textbf{9.93} &         9.19  &
                \multicolumn{11}{c|}{\multirow{2}{*}{}}  \\
        \hline
    
        \end{tabular}
        \addtolength{\tabcolsep}{5pt}

%% file: arxiv.bbl
\begin{thebibliography}{10}\itemsep=-1pt

\bibitem{Artetxe18Multilingual}
Mikel Artetxe and Holger Schwenk.
\newblock Massively multilingual sentence embeddings for zero-shot
  cross-lingual transfer and beyond.
\newblock {\em CoRR}, abs/1812.10464, 2018.

\bibitem{bay2006surf}
Herbert Bay, Tinne Tuytelaars, and Luc Van~Gool.
\newblock Surf: Speeded up robust features.
\newblock In {\em European conference on computer vision}, pages 404--417.
  Springer, 2006.

\bibitem{BrislinCrossCultural}
Richard~W. Brislin.
\newblock Back-translation for cross-cultural research.
\newblock {\em Journal of Cross-Cultural Psychology}, 1(3):185--216, 1970.

\bibitem{chalapathy2018anomaly}
Raghavendra Chalapathy, Aditya~Krishna Menon, and Sanjay Chawla.
\newblock Anomaly detection using one-class neural networks.
\newblock {\em arXiv preprint arXiv:1802.06360}, 2018.

\bibitem{choi2018stargan}
Yunjey Choi, Minje Choi, Munyoung Kim, Jung-Woo Ha, Sunghun Kim, and Jaegul
  Choo.
\newblock Stargan: Unified generative adversarial networks for multi-domain
  image-to-image translation.
\newblock In {\em Proceedings of the IEEE Conference on Computer Vision and
  Pattern Recognition}, pages 8789--8797, 2018.

\bibitem{cosmo2017consistent}
Luca Cosmo, Emanuele Rodola, Andrea Albarelli, Facundo M{\'e}moli, and Daniel
  Cremers.
\newblock Consistent partial matching of shape collections via sparse modeling.
\newblock In {\em Computer Graphics Forum}, volume~36, pages 209--221. Wiley
  Online Library, 2017.

\bibitem{DwibediTimeConsistency19}
Debidatta Dwibedi, Yusuf Aytar, Jonathan Tompson, Pierre Sermanet, and Andrew
  Zisserman.
\newblock Temporal cycle-consistency learning.
\newblock {\em CoRR}, abs/1904.07846, 2019.

\bibitem{backtranslation2018}
Sergey Edunov, Myle Ott, Michael Auli, and David Grangier.
\newblock Understanding back-translation at scale.
\newblock {\em CoRR}, abs/1808.09381, 2018.

\bibitem{garg2016unsupervised}
Ravi Garg, Vijay~Kumar BG, Gustavo Carneiro, and Ian Reid.
\newblock Unsupervised cnn for single view depth estimation: Geometry to the
  rescue.
\newblock In {\em European Conference on Computer Vision}, pages 740--756.
  Springer, 2016.

\bibitem{godard2017unsupervised}
Cl{\'e}ment Godard, Oisin Mac~Aodha, and Gabriel~J Brostow.
\newblock Unsupervised monocular depth estimation with left-right consistency.
\newblock In {\em Proceedings of the IEEE Conference on Computer Vision and
  Pattern Recognition}, pages 270--279, 2017.

\bibitem{guillemin1974differential}
V. Guillemin and A. Pollack.
\newblock {\em Differential Topology}.
\newblock Mathematics Series. Prentice-Hall, 1974.

\bibitem{guo2017calibration}
Chuan Guo, Geoff Pleiss, Yu Sun, and Kilian~Q. Weinberger.
\newblock On calibration of modern neural networks, 2017.

\bibitem{resnet}
Kaiming He, Xiangyu Zhang, Shaoqing Ren, and Jian Sun.
\newblock Deep residual learning for image recognition.
\newblock {\em CoRR}, abs/1512.03385, 2015.

\bibitem{hertzmann2001image}
Aaron Hertzmann, Charles~E Jacobs, Nuria Oliver, Brian Curless, and David~H
  Salesin.
\newblock Image analogies.
\newblock In {\em Proceedings of the 28th annual conference on Computer
  graphics and interactive techniques}, pages 327--340. ACM, 2001.

\bibitem{hickson2019floors}
Steven Hickson, Karthik Raveendran, Alireza Fathi, Kevin Murphy, and Irfan
  Essa.
\newblock Floors are flat: Leveraging semantics for real-time surface normal
  prediction.
\newblock In {\em Proceedings of the IEEE International Conference on Computer
  Vision Workshops}, pages 0--0, 2019.

\bibitem{HoffmancyCADA17}
Judy Hoffman, Eric Tzeng, Taesung Park, Jun{-}Yan Zhu, Phillip Isola, Kate
  Saenko, Alexei~A. Efros, and Trevor Darrell.
\newblock Cycada: Cycle-consistent adversarial domain adaptation.
\newblock {\em CoRR}, abs/1711.03213, 2017.

\bibitem{Huang:2013:CSM:2600289.2600314}
Qi-Xing Huang and Leonidas Guibas.
\newblock Consistent shape maps via semidefinite programming.
\newblock In {\em Proceedings of the Eleventh Eurographics/ACMSIGGRAPH
  Symposium on Geometry Processing}, SGP '13, pages 177--186, Aire-la-Ville,
  Switzerland, Switzerland, 2013. Eurographics Association.

\bibitem{appoloscape}
Xinyu Huang, Xinjing Cheng, Qichuan Geng, Binbin Cao, Dingfu Zhou, Peng Wang,
  Yuanqing Lin, and Ruigang Yang.
\newblock The apolloscape dataset for autonomous driving.
\newblock {\em CoRR}, abs/1803.06184, 2018.

\bibitem{pix2pix}
Phillip Isola, Jun{-}Yan Zhu, Tinghui Zhou, and Alexei~A. Efros.
\newblock Image-to-image translation with conditional adversarial networks.
\newblock {\em CoRR}, abs/1611.07004, 2016.

\bibitem{jo2017measuring}
Jason Jo and Yoshua Bengio.
\newblock Measuring the tendency of cnns to learn surface statistical
  regularities.
\newblock {\em arXiv preprint arXiv:1711.11561}, 2017.

\bibitem{DBLP:journals/corr/JohnsonAL16}
Justin Johnson, Alexandre Alahi, and Fei{-}Fei Li.
\newblock Perceptual losses for real-time style transfer and super-resolution.
\newblock {\em CoRR}, abs/1603.08155, 2016.

\bibitem{Kochenderfer:2015:DMU:2815660}
Mykel~J. Kochenderfer, Christopher Amato, Girish Chowdhary, Jonathan~P. How,
  Hayley J.~Davison Reynolds, Jason~R. Thornton, Pedro~A. Torres-Carrasquillo,
  N.~Kemal \"{U}re, and John Vian.
\newblock {\em Decision Making Under Uncertainty: Theory and Application}.
\newblock The MIT Press, 1st edition, 2015.

\bibitem{Kokkinos16}
Iasonas Kokkinos.
\newblock Ubernet: Training a 'universal' convolutional neural network for
  low-, mid-, and high-level vision using diverse datasets and limited memory.
\newblock {\em CoRR}, abs/1609.02132, 2016.

\bibitem{Koller:2009:PGM:1795555}
Daphne Koller and Nir Friedman.
\newblock {\em Probabilistic Graphical Models: Principles and Techniques -
  Adaptive Computation and Machine Learning}.
\newblock The MIT Press, 2009.

\bibitem{kusupati2019normal}
Uday Kusupati, Shuo Cheng, Rui Chen, and Hao Su.
\newblock Normal assisted stereo depth estimation.
\newblock {\em arXiv preprint arXiv:1911.10444}, 2019.

\bibitem{lakshminarayanan2017simple}
Balaji Lakshminarayanan, Alexander Pritzel, and Charles Blundell.
\newblock Simple and scalable predictive uncertainty estimation using deep
  ensembles.
\newblock In {\em Advances in Neural Information Processing Systems}, pages
  6402--6413, 2017.

\bibitem{Lample19crosslingual}
Guillaume Lample and Alexis Conneau.
\newblock Cross-lingual language model pretraining.
\newblock {\em CoRR}, abs/1901.07291, 2019.

\bibitem{lecun2006tutorial}
Yann LeCun, Sumit Chopra, and Raia Hadsell.
\newblock A tutorial on energy-based learning.
\newblock 2006.

\bibitem{learningWithoutForgetting}
Zhizhong Li and Derek Hoiem.
\newblock Learning without forgetting.
\newblock {\em CoRR}, abs/1606.09282, 2016.

\bibitem{researchdoom}
Aravindh Mahendran, Hakan Bilen, Jo{\~{a}}o~F. Henriques, and Andrea Vedaldi.
\newblock Researchdoom and cocodoom: Learning computer vision with games.
\newblock {\em CoRR}, abs/1610.02431, 2016.

\bibitem{mirza2014conditional}
Mehdi Mirza and Simon Osindero.
\newblock Conditional generative adversarial nets.
\newblock {\em arXiv preprint arXiv:1411.1784}, 2014.

\bibitem{ovadia2019trust}
Yaniv Ovadia, Emily Fertig, Jie Ren, Zachary Nado, D Sculley, Sebastian
  Nowozin, Joshua~V. Dillon, Balaji Lakshminarayanan, and Jasper Snoek.
\newblock Can you trust your model's uncertainty? evaluating predictive
  uncertainty under dataset shift, 2019.

\bibitem{Ovsjanikov:2012:FMF:2185520.2185526}
Maks Ovsjanikov, Mirela Ben-Chen, Justin Solomon, Adrian Butscher, and Leonidas
  Guibas.
\newblock Functional maps: A flexible representation of maps between shapes.
\newblock {\em ACM Trans. Graph.}, 31(4):30:1--30:11, July 2012.

\bibitem{proctor2017tolerances}
F Proctor, Marek Franaszek, and J Michaloski.
\newblock Tolerances and uncertainty in robotic systems.
\newblock In {\em ASME 2017 International Mechanical Engineering Congress and
  Exposition}. American Society of Mechanical Engineers Digital Collection,
  2017.

\bibitem{Geonet18}
Xiaojuan Qi, Renjie Liao, Zhengzhe Liu, Raquel Urtasun, and Jiaya Jia.
\newblock Geonet: Geometric neural network for joint depth and surface normal
  estimation.
\newblock In {\em Proceedings of the IEEE Conference on Computer Vision and
  Pattern Recognition}, pages 283--291, 2018.

\bibitem{reddi2019convergence}
Sashank~J Reddi, Satyen Kale, and Sanjiv Kumar.
\newblock On the convergence of adam and beyond.
\newblock {\em arXiv preprint arXiv:1904.09237}, 2019.

\bibitem{RonnebergerFB15}
Olaf Ronneberger, Philipp Fischer, and Thomas Brox.
\newblock U-net: Convolutional networks for biomedical image segmentation.
\newblock {\em CoRR}, abs/1505.04597, 2015.

\bibitem{sener2017active}
Ozan Sener and Silvio Savarese.
\newblock Active learning for convolutional neural networks: A core-set
  approach.
\newblock {\em arXiv preprint arXiv:1708.00489}, 2017.

\bibitem{sharif2014cnn}
Ali Sharif~Razavian, Hossein Azizpour, Josephine Sullivan, and Stefan Carlsson.
\newblock Cnn features off-the-shelf: an astounding baseline for recognition.
\newblock In {\em Proceedings of the IEEE conference on computer vision and
  pattern recognition workshops}, pages 806--813, 2014.

\bibitem{Silberman2012}
Nathan Silberman, Derek Hoiem, Pushmeet Kohli, and Rob Fergus.
\newblock {\em Indoor Segmentation and Support Inference from RGBD Images},
  pages 746--760.
\newblock Springer Berlin Heidelberg, Berlin, Heidelberg, 2012.

\bibitem{standley2019}
Trevor {Standley}, Amir~R. {Zamir}, Dawn {Chen}, Leonidas {Guibas}, Jitendra
  {Malik}, and Silvio {Savarese}.
\newblock {Which Tasks Should Be Learned Together in Multi-task Learning?}
\newblock In {\em International Conference on Machine Learning}, 2020.

\bibitem{stewart2012essential}
James Stewart.
\newblock {\em Essential calculus: Early transcendentals}.
\newblock Cengage Learning, 2012.

\bibitem{straub2019replica}
Julian Straub, Thomas Whelan, Lingni Ma, Yufan Chen, Erik Wijmans, Simon Green,
  Jakob~J Engel, Raul Mur-Artal, Carl Ren, Shobhit Verma, et~al.
\newblock The replica dataset: A digital replica of indoor spaces.
\newblock {\em arXiv preprint arXiv:1906.05797}, 2019.

\bibitem{WangCycleConsistency19}
Xiaolong Wang, Allan Jabri, and Alexei~A. Efros.
\newblock Learning correspondence from the cycle-consistency of time.
\newblock {\em CoRR}, abs/1903.07593, 2019.

\bibitem{wu2018group}
Yuxin Wu and Kaiming He.
\newblock Group normalization.
\newblock In {\em Proceedings of the European Conference on Computer Vision
  (ECCV)}, pages 3--19, 2018.

\bibitem{xu2018pad}
Dan Xu, Wanli Ouyang, Xiaogang Wang, and Nicu Sebe.
\newblock Pad-net: Multi-tasks guided prediction-and-distillation network for
  simultaneous depth estimation and scene parsing.
\newblock In {\em Proceedings of the IEEE Conference on Computer Vision and
  Pattern Recognition}, pages 675--684, 2018.

\bibitem{yin2019enforcing}
Wei Yin, Yifan Liu, Chunhua Shen, and Youliang Yan.
\newblock Enforcing geometric constraints of virtual normal for depth
  prediction.
\newblock In {\em Proceedings of the IEEE International Conference on Computer
  Vision}, pages 5684--5693, 2019.

\bibitem{taskonomy18}
Amir~R Zamir, Alexander Sax, William Shen, Leonidas~J Guibas, Jitendra Malik,
  and Silvio Savarese.
\newblock Taskonomy: Disentangling task transfer learning.
\newblock In {\em Proceedings of the IEEE Conference on Computer Vision and
  Pattern Recognition}, pages 3712--3722, 2018.

\bibitem{zhang2017physically}
Yinda Zhang, Shuran Song, Ersin Yumer, Manolis Savva, Joon-Young Lee, Hailin
  Jin, and Thomas Funkhouser.
\newblock Physically-based rendering for indoor scene understanding using
  convolutional neural networks.
\newblock In {\em Proceedings of the IEEE Conference on Computer Vision and
  Pattern Recognition}, pages 5287--5295, 2017.

\bibitem{Zhang17MTL}
Yu Zhang and Qiang Yang.
\newblock A survey on multi-task learning.
\newblock {\em CoRR}, abs/1707.08114, 2017.

\bibitem{Zhang18PathInvariant}
Zaiwei Zhang, Zhenxiao Liang, Lemeng Wu, Xiaowei Zhou, and Qixing Huang.
\newblock Path-invariant map networks.
\newblock {\em CoRR}, abs/1812.11647, 2018.

\bibitem{ZhouKAHE16}
Tinghui Zhou, Philipp Kr{\"{a}}henb{\"{u}}hl, Mathieu Aubry, Qixing Huang, and
  Alexei~A. Efros.
\newblock Learning dense correspondence via 3d-guided cycle consistency.
\newblock {\em CoRR}, abs/1604.05383, 2016.

\bibitem{cycleGan17}
Jun{-}Yan Zhu, Taesung Park, Phillip Isola, and Alexei~A. Efros.
\newblock Unpaired image-to-image translation using cycle-consistent
  adversarial networks.
\newblock {\em CoRR}, abs/1703.10593, 2017.

\bibitem{zou2018df}
Yuliang Zou, Zelun Luo, and Jia-Bin Huang.
\newblock Df-net: Unsupervised joint learning of depth and flow using
  cross-task consistency.
\newblock In {\em Proceedings of the European Conference on Computer Vision
  (ECCV)}, pages 36--53, 2018.

\end{thebibliography}
